\documentclass{article} 
\usepackage{iclr2027_conference,times}

\usepackage{amsmath,amsfonts,bm}

\def\eqref#1{equation~\ref{#1}}

\def\1{\bm{1}}

\DeclareMathAlphabet{\mathsfit}{\encodingdefault}{\sfdefault}{m}{sl}
\SetMathAlphabet{\mathsfit}{bold}{\encodingdefault}{\sfdefault}{bx}{n}

\usepackage{hyperref}
\hypersetup{hidelinks}
\usepackage{url}
\usepackage{booktabs}
\usepackage{longtable}
\usepackage{amsmath}
\usepackage{amssymb}
\usepackage{amsthm}
\usepackage{graphicx}
\usepackage{multirow,multicol}
\usepackage{tabularx}
\usepackage{float}
\usepackage{wrapfig}
\usepackage{cleveref}
\usepackage{adjustbox}
\usepackage{comment}
\usepackage{enumitem}
\usepackage{xcolor}
\usepackage{soul}
\usepackage{caption}

\graphicspath{{figures/}}

\newcommand{\OneBitMicroAurocSht}{96.32\%}
\newcommand{\OneBitMicroAurocXd}{84.93\%}
\newcommand{\OneBitMicroAurocUcf}{81.57\%}

\newcommand{\auroc}{\mathrm{AUROC}}
\newcommand{\aurocmicro}{\mathrm{Micro\text{-}AUROC}}
\newcommand{\aurocwithin}{\mathrm{Within\text{-}AUROC}}
\newcommand{\auroccross}{\mathrm{Cross\text{-}AUROC}}
\newcommand{\neff}{N_{\mathrm{eff}}}

\newtheorem{proposition}{Proposition}
\newtheorem{corollary}{Corollary}

\definecolor{withinhl}{HTML}{FFDC96}
\definecolor{crosshl}{HTML}{CEE1FE}

\title{
Look Inside Each Video: Rethinking \\ How Video Anomaly Detection Is Evaluated
}

\author{
Inpyo Song \\
SungKyunKwan University \\
South Korea \\
\texttt{songinpyo@skku.edu}
\And
Jangwon Lee \\
SungKyunKwan University \\
South Korea \\
\texttt{leejang@skku.edu}
}

\newif\ifarxiv
\arxivtrue          

\ifarxiv
  \iclrfinalcopy    
\fi

\begin{document}

\maketitle

\ifarxiv
  \fancyhead{}      
\fi

\maketitle

\begin{abstract}
Video anomaly detection (VAD) aims to localize anomalous events by identifying when they occur.
A common evaluation pools frame-level anomaly scores across all test videos and measures whether anomalous frames receive higher scores than normal frames (Micro-AUROC).
This pooling creates comparisons both within the same video and across different videos.
However, the pooled evaluation protocol commonly used in current VAD benchmarks makes Micro-AUROC heavily dependent on cross-video comparisons, which do not directly test whether anomalous frames rank above normal frames within the same video.
To quantify this imbalance, we decompose Micro-AUROC into within-video and cross-video comparisons.
Across four widely used weakly supervised VAD benchmarks, within-video pairs account for only $0.071$--$0.388\%$ of all anomalous--normal frame pairs.
This imbalance can make Micro-AUROC a poor indicator of within-video ranking.
Video-level classifiers that assign a single score to each video achieve $81.40$--$97.18\%$ Micro-AUROC while obtaining $50\%$ Within-AUROC.
For existing weakly supervised VAD detectors, 
replacing all frame scores in each video with their mean sets Within-AUROC to $50\%$.
Across 120 runs, 
this replacement retains a median $98.8\%$ of the original Micro-AUROC margin above chance.
Because the decomposition depends on the evaluation protocol rather than the training supervision, 
the same distinction applies whenever VAD frame scores are pooled across videos.
High Micro-AUROC alone therefore does not establish temporal localization. Our analysis reveals a limitation of pooled VAD evaluation and motivates reporting within-video ranking alongside Micro-AUROC.
\end{abstract}

\section{Introduction}
\begin{figure}[t]
\centering
\includegraphics[width=\linewidth]{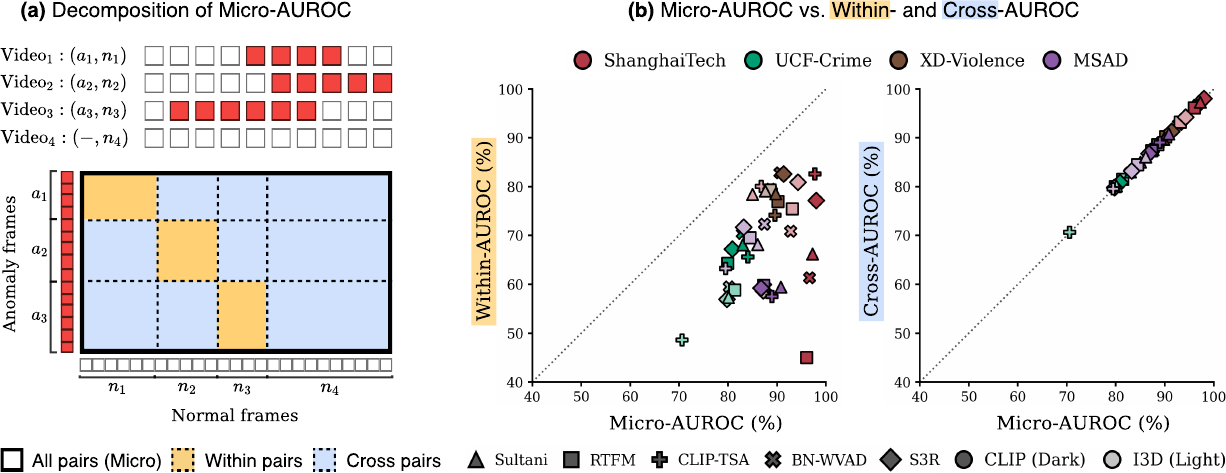}
\vspace{-1em}
\caption{
\textbf{Micro-AUROC closely matches Cross-AUROC.}
(a) Anomalous--normal frame pairs are divided into within-video
and cross-video comparisons.
(b) Micro-AUROC is compared with Within-AUROC and Cross-AUROC
across four benchmarks and five WSVAD methods.
Detectors with similar Micro-AUROC can have substantially different Within-AUROC.
}
\label{fig:metric-anatomy}
\end{figure}

Weakly supervised video anomaly detection (WSVAD) uses video-level supervision to localize anomalous events by
identifying when they occur~\citep{s3r,song2025anomaly}.
At test time, WSVAD methods assign an anomaly score to each frame~\citep{rtfm}.
Micro-AUROC evaluates these scores by pooling all frames across the test set, whereas Macro-AUROC averages per-video AUROC over videos containing both anomalous and normal frames.
Despite this distinction, WSVAD benchmarks predominantly report pooled metrics, such as Micro-AUROC and Average Precision (AP)~\citep{sultani2018real,wu2020not}.
By pooling frames across videos, these metrics assess how anomalous and normal frames are ranked across the test set rather than specifically within individual videos.
We examine what this pooled evaluation captures by decomposing the anomalous–normal frame pairs underlying Micro-AUROC into two groups (\Cref{fig:metric-anatomy}(a)): within-video pairs and cross-video pairs.
Within-video pairs directly test whether anomalous frames rank above normal frames in the same video.
In contrast, cross-video pairs compare frames from different videos and may be ranked correctly even when a scorer fails to distinguish anomalous from normal frames within a video.

This distinction raises a basic question:
\emph{how much does Micro-AUROC depend on distinguishing anomalous from normal frames within the same video?}
We answer this question in three steps.

First, we decompose Micro-AUROC into \emph{Within-AUROC}, computed
over within-video pairs, and \emph{Cross-AUROC}, computed over
cross-video pairs.
Across four widely used WSVAD benchmarks
(ShanghaiTech, XD-Violence, UCF-Crime, and MSAD)
within-video pairs account for only $0.071\%$--$0.388\%$ of all anomalous--normal frame pairs.
Because within-video pairs constitute such a small fraction of all pairs, even increasing Within-AUROC from chance to perfection can raise Micro-AUROC by at most $0.036$--$0.194$ percentage points, with Cross-AUROC held fixed.
Consistent with this pair-count analysis,
\Cref{fig:metric-anatomy}(b) shows that detector-level Micro-AUROC closely tracks Cross-AUROC,
while Within-AUROC can differ substantially.
We further derive an upper bound on the within-video pair fraction
that decreases inversely with the number of videos when frames
of each class remain broadly distributed across videos.
We call this effect \emph{temporal dilution}.

Second, this is not merely a theoretical possibility.
We train video-level classifiers using only video-level labels
and assign each video’s predicted score to all of its frames.
All video-constant outputs have Within-AUROC of $50\%$ by construction.
The best classifier achieves a Micro-AUROC of $81.40\%$ to $97.18\%$ across the benchmarks.
Thus, high Micro-AUROC does not necessarily indicate that a model distinguishes anomalous from normal frames within individual videos.

Finally, we test whether existing detectors retain high Micro-AUROC after removing their within-video ranking.
We replace every frame score in a video with that video's mean,
making all anomalous--normal frame pairs within the video tied.
If Micro-AUROC substantially depended on within-video ranking,
this replacement should cause a substantial decrease.
Instead, across 120 reproduced WSVAD runs, \emph{video-mean replacement}
preserves a median $98.8\%$ of the original Micro-AUROC margin above chance.

Taken together, these results answer our question: on the benchmarks studied, Micro-AUROC depends only weakly on distinguishing anomalous from normal frames within the same video.
High Micro-AUROC therefore does not by itself establish temporal localization.
The same decomposition applies whenever VAD frame scores are pooled across videos for Micro-AUROC evaluation.
We therefore argue that evidence of temporal localization must come from evaluating within-video ranking, not from pooled metrics alone.

\section{Related work}
\label{sec:related-work}

\paragraph{Video anomaly detection.}
Video anomaly detection assigns anomaly scores over time to identify unusual events in video.
Existing methods are commonly studied under different supervision settings,
including semi-supervised, weakly supervised, and training-free VAD,
with different benchmarks and training assumptions~\citep{rtfm,lavad,song2025anomaly,li2026vadtree,song2026bounding}.
Despite these differences, these settings share a common evaluation target: 
a detector assigns anomaly scores over time, 
and performance is commonly measured 
by how these scores rank anomalous frames above normal frames (\Cref{app:metric-survey}).
Our analysis concerns this shared evaluation step rather than a particular training paradigm.
We use WSVAD as our primary empirical setting because its standard evaluation pools frame scores across many videos, 
making it a direct test of our question.
We then extend the analysis to semi-supervised and training-free VAD.

\paragraph{Evaluation in video anomaly detection.}
Frame-level VAD evaluation typically pools predictions across the
test set and summarizes how well anomaly scores separate anomalous
from normal frames using AUROC or AP
~\citep{li2013anomaly,lu2013abnormal,liu2018future,
sultani2018real,wu2020not,msad2024}.
Subsequent work has changed different parts of this evaluation.
One approach removes all-normal videos and evaluates only
anomaly-containing videos, reducing the influence of predictions on
videos that contain no anomaly ~\citep{lv2021localizing,lv2023unbiased}.
Another computes AUROC separately for each mixed-label video and
averages across videos, thereby removing comparisons between frames
from different videos ~\citep{acsintoae2022ubnormal,ristea2024selfdistilled}.
Complementary evaluations ask different localization questions.
Temporal detection mAP evaluates predicted event intervals and, in
fine-grained settings, anomaly categories ~\citep{wu2024vadclip}, 
while fixed-operating-point evaluation measures detection performance at specified false-positive rates rather than averaging over all thresholds ~\citep{acharya2026road}.
These alternatives change the evaluated videos, the aggregation unit,
the localization target, or the operating point.
Our analysis instead leaves pooled Micro-AUROC unchanged and asks how
its anomalous--normal comparisons are divided between frames from the
same video and frames from different videos.

\paragraph{Decomposing pooled AUROC.}
The pairwise view of AUROC makes it possible to separate an aggregate ranking according to group membership.
Prior work in fairness and bipartite ranking uses this view to distinguish within-group from cross-group positive--negative
comparisons and to show that their relative weights depend on group and class composition~\citep{borkan2019limitations,kallus2019fairness,vogel2021learning}.
A related analysis in open-world VAD separates comparisons according
to the queried anomaly definition, showing that one type of
comparison can dominate the reported score
~\citep{song2026rethinking}.
We instead partition pairs by video identity.
With this grouping, within-group comparisons are exactly
anomalous--normal pairs from the same video, while cross-group
comparisons come from different videos.
This lets us quantify how benchmark composition determines the
direct weight of within-video ranking in Micro-AUROC and characterize
how high Micro-AUROC can remain when every frame in a video receives
the same score.

\section{What does Micro-AUROC measure?}
\label{sec:decomposition}

\subsection{Within-video and cross-video comparisons}

Micro-AUROC is computed after pooling the frame scores from all test videos.
Equivalently, it measures how often an anomalous frame is ranked above a
normal frame, with tied pairs receiving half credit~\citep{song2026vlm}.
Each anomalous--normal pair either comes from the same video or from two
different videos.
We separate Micro-AUROC according to these two pair populations.

Let video $i$ contain $a_i$ anomalous frames and $n_i$ normal frames, and let
\begin{equation}
A = \sum_i a_i,
\qquad
N = \sum_i n_i,
\qquad
W = \sum_i a_i n_i.
\end{equation}
The pooled evaluation contains $AN$ anomalous--normal pairs in total, of
which $W$ come from the same video.
We define \emph{Within-AUROC} as the average pairwise correctness over these
$W$ within-video pairs and \emph{Cross-AUROC} as the average over the remaining
$AN-W$ cross-video pairs.

\begin{proposition}[Exact decomposition by video identity]
\label{prop:decomp}
When both pair populations are nonempty,
\begin{equation}
\aurocmicro
=
w\,\aurocwithin
+
(1-w)\,\auroccross,
\qquad
w = \frac{W}{AN}.
\label{eq:decomp}
\end{equation}
\end{proposition}

The weight $w$ depends only on the evaluated frame labels and video
boundaries, not on the detector scores.
With AUROC on the $[0,1]$ scale, \Cref{eq:decomp} gives
\begin{equation}
\left|\aurocmicro-\auroccross\right|
= w\left|\aurocwithin-\auroccross\right|
\leq w.
\label{eq:decomp_cap}
\end{equation}
Micro-AUROC must therefore be close to Cross-AUROC when $w$ is
small, regardless of the detector's within-video ranking.
Appendix~\ref{app:proofs} proves the decomposition by partitioning
anomalous--normal pairs according to video identity.

\subsection{Temporal dilution}

The within-video pair fraction becomes small when anomalous and normal
frames are spread across many videos.
To make this precise, define
\begin{equation}
p_i^+ = \frac{a_i}{A},
\qquad
p_i^- = \frac{n_i}{N},
\qquad
w = \sum_i p_i^+ p_i^-.
\label{eq:dilution-prob}
\end{equation}
Here, $w$ is the probability that a randomly sampled anomalous frame and
normal frame come from the same video.

This probability satisfies
\begin{equation}
w
\leq
\min\left\{
\max_i p_i^+,
\max_i p_i^-
\right\}.
\label{eq:dilution}
\end{equation}
If the largest fraction of anomalous frames and the largest fraction of
normal frames contributed by any single video both scale as $O(1/V)$,
where $V$ is the number of videos, then $w=O(1/V)$.
We call this decrease in the within-video pair fraction
\emph{temporal dilution}.
The number of videos alone does not determine $w$, because the distribution
of each class across videos also matters.

\begin{corollary}[Maximum direct contribution of Within-AUROC]
\label{cor:maximum}
If Cross-AUROC is fixed, improving Within-AUROC from $50\%$ to $100\%$
can increase Micro-AUROC by at most $w/2$.
\end{corollary}

\begin{wraptable}{r}{0.43\textwidth}
\vspace{-1.4em}
\centering
\small
\setlength{\tabcolsep}{2.8pt}
\renewcommand{\arraystretch}{1.08}

\caption{
Within-video pair fraction and maximum direct contribution to Micro-AUROC.
}
\label{tab:budget}

\vspace{0.3em}

\begin{tabular}{lccr}
\toprule
Benchmark &
$V$ &
$w$ &
\shortstack{Max. gain (pp)} \\
\midrule
ShanghaiTech & 199 & $0.140\%$ & $0.070$ \\
UCF-Crime    & 290 & $0.388\%$ & $0.194$ \\
XD-Violence  & 800 & $0.071\%$ & $0.036$ \\
MSAD         & 240 & $0.206\%$ & $0.103$ \\
\bottomrule
\end{tabular}

\vspace{-1.3em}
\end{wraptable}

\Cref{tab:budget} shows that only $0.071\%$--$0.388\%$ of
anomalous--normal pairs come from the same video on the four WSVAD benchmarks.
Holding Cross-AUROC fixed, improving Within-AUROC from $50\%$ to $100\%$
can therefore increase Micro-AUROC by at most
$0.036$--$0.194$ percentage points.
With such small within-video pair fractions, Micro-AUROC is numerically
close to Cross-AUROC, as shown in \Cref{fig:metric-anatomy}(b).
The same figure shows that detectors with similar Micro-AUROC can nevertheless
have substantially different Within-AUROC.

\begin{wrapfigure}{r}{0.43\textwidth}
\vspace{-1.3em}
\centering

\includegraphics[
width=\linewidth
]{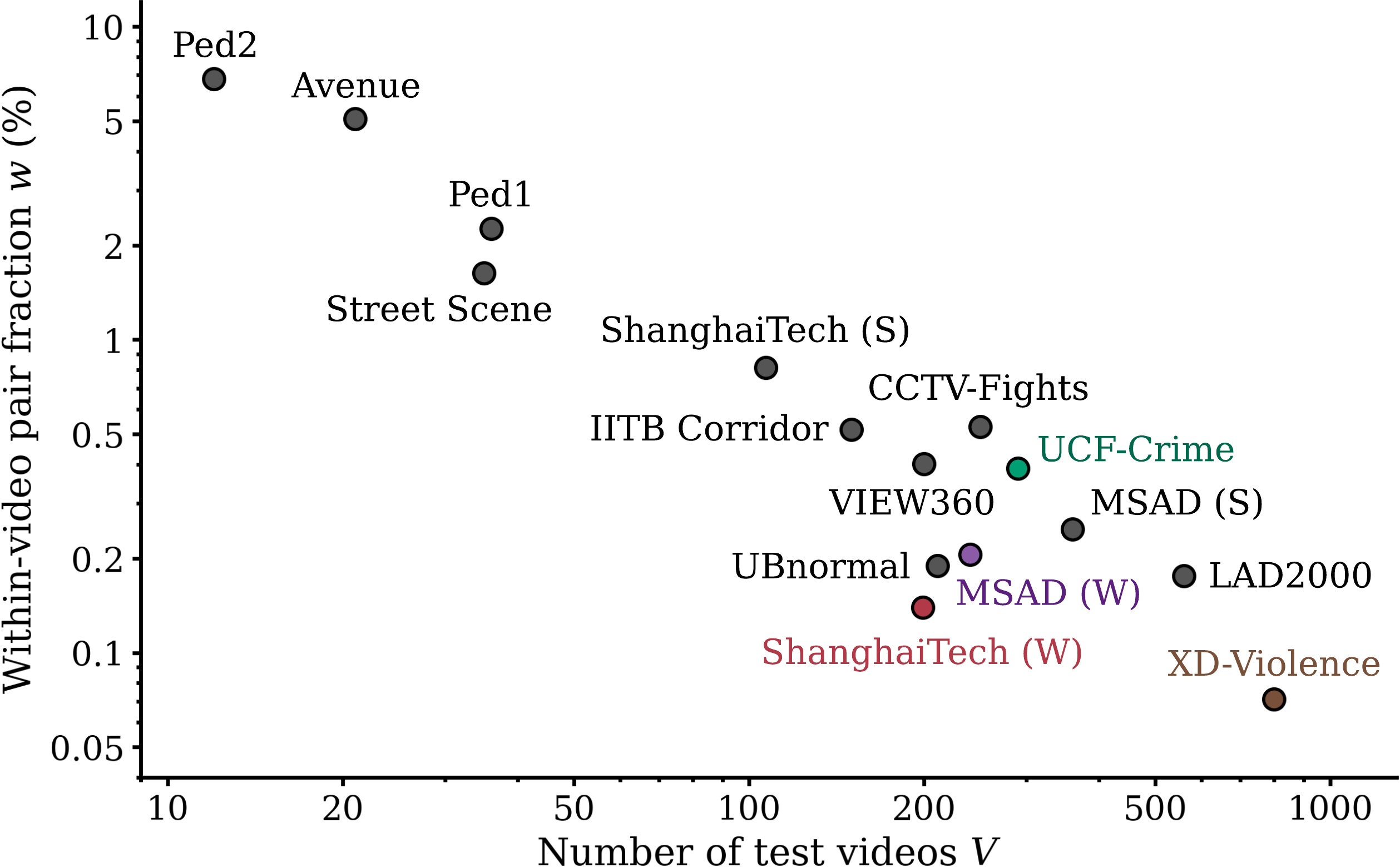}

\caption{
Within-video pair fractions across VAD benchmarks.
Colored points denote the four WSVAD benchmarks used in our main experiments.
}
\label{fig:vad-temporal-dilution}

\vspace{-2.5em}
\end{wrapfigure}

The four WSVAD benchmarks are not isolated cases.
\Cref{fig:vad-temporal-dilution} shows the within-video pair fraction across
a broader set of 13 VAD benchmarks.
Appendix~\ref{app:benchmark-composition} reports the statistics for each
test split.
UCSD Ped2 and Avenue have within-video fractions of several percent,
whereas UBnormal and XD-Violence fall below $0.2\%$.
The relationship is not determined by the number of videos alone, since
the distribution of anomalous and normal frames also affects $w$.
The broader pattern is consistent with temporal dilution.

The decomposition therefore shows that pooled Micro-AUROC can assign very
little weight to within-video comparisons, 
both on the WSVAD benchmarks 
and more broadly across VAD benchmarks.

\begin{table}[t]
\centering
\caption{
Micro-AUROC (\%) from video-level anomaly classifiers.
Each classifier predicts a single anomaly score per video and assigns that score to all frames in the video.
Results are averaged over three runs.
The oracle uses frame-level test labels to order videos by their
ground-truth fraction of anomalous frames.
Within-AUROC is $50\%$ for every row.
}
\label{tab:video-classifiers}
\small
\renewcommand{\arraystretch}{1.2}
\begin{tabular*}{\linewidth}{@{\extracolsep{\fill}}llrrrr@{}}
\toprule
Classifier & Feature & ShanghaiTech & UCF-Crime & XD-Violence & MSAD \\
\midrule
(a) Mean-pool        & CLIP & 96.87$\pm0.00$ & 78.50$\pm0.04$ & 88.76$\pm0.02$ & \textbf{89.76}$\pm0.01$ \\
(b) Sequence-linear  & CLIP & 96.60$\pm0.01$ & 80.03$\pm0.21$ & \textbf{90.78}$\pm0.04$ & 89.34$\pm0.02$ \\
(c) Bi-GRU           & CLIP & 97.16$\pm0.01$ & 77.84$\pm0.92$ & 89.92$\pm0.20$ & 88.99$\pm0.50$ \\
(d) Transformer      & CLIP & \textbf{97.18}$\pm0.15$ & 80.40$\pm0.87$ & 88.13$\pm1.77$ & 88.82$\pm0.27$ \\
\midrule
(a) Mean-pool        & I3D  & 92.06$\pm0.04$ & 79.21$\pm0.10$ & 89.22$\pm0.02$ & 86.20$\pm0.05$ \\
(b) Sequence-linear  & I3D  & 91.18$\pm0.07$ & \textbf{81.40}$\pm0.06$ & 89.33$\pm0.20$ & 86.47$\pm0.08$ \\
(c) Bi-GRU           & I3D  & 91.10$\pm0.48$ & 77.26$\pm0.09$ & 90.45$\pm0.64$ & 84.63$\pm0.28$ \\
(d) Transformer      & I3D  & 85.34$\pm2.12$ & 79.07$\pm0.77$ & 87.30$\pm1.42$ & 85.17$\pm0.81$ \\
\midrule
\multicolumn{2}{@{}l}{Oracle using test labels}
& 98.35 & 92.95 & 95.46 & 96.90 \\
\bottomrule
\end{tabular*}
\end{table}

\section{Video-Level Anomaly Classification}
\label{sec:video-classification}

The previous section shows that only a small fraction of the comparisons
in Micro-AUROC involve anomalous and normal frames from the same video.
This raises a natural question:
can a video-level anomaly classifier achieve high Micro-AUROC without
distinguishing anomalous from normal frames within the same video?
We test the simplest case by assigning one anomaly score to each video
and repeating that score across all of its frames.
Every within-video anomalous--normal pair is then tied, so Within-AUROC is
$50\%$ by construction.

\subsection{Maximum Micro-AUROC}

We first characterize the maximum Micro-AUROC possible when every frame in
a video receives the same anomaly score.
In this setting, the ordering between videos determines all cross-video
comparisons.
Consider two videos $i$ and $j$.
Assigning a higher anomaly score to video $i$ correctly orders
$a_i n_j$ cross-video pairs, while assigning a higher score to video $j$
correctly orders $a_j n_i$ pairs.
The first ordering is preferable exactly when
\begin{equation}
a_i n_j \geq a_j n_i
\qquad\Longleftrightarrow\qquad
\frac{a_i}{a_i+n_i}
\geq
\frac{a_j}{a_j+n_j}.
\end{equation}

\begin{proposition}[Optimal ordering with one anomaly score per video]
\label{prop:video-score-optimum}
If every frame in a video receives the same anomaly score,
Micro-AUROC is maximized by ordering videos according to their
ground-truth fraction of anomalous frames $a_i/(a_i+n_i)$,
with arbitrary tie breaking among equal fractions.
\end{proposition}

The pairwise preferences above are transitive, so ordering videos by their
fraction of anomalous frames maximizes the number of correctly ordered
cross-video pairs.
Within-AUROC remains fixed at $50\%$, so the same ordering maximizes
Micro-AUROC by \Cref{eq:decomp}.
This ordering is also consistent with likelihood-ratio ordering in
bipartite ranking and ROC analysis
~\citep{fawcett2007pav,clemenccon2008ranking,menon2016bipartite}.
Appendix~\ref{app:proofs} gives the proof.

Using the frame-level test labels to construct this ordering gives
$98.35\%$ Micro-AUROC on ShanghaiTech, $92.95\%$ on UCF-Crime,
$95.46\%$ on XD-Violence, and $96.90\%$ on MSAD
(\Cref{tab:video-classifiers}).
Thus, assigning only one anomaly score to each video can yield
$92.95\%$--$98.35\%$ Micro-AUROC on these benchmarks,
even though Within-AUROC is $50\%$.

\subsection{Training Video-Level Anomaly Classifiers}

\begin{wrapfigure}{r}{0.25\textwidth}
    \centering
    \vspace{-4.3em}
    \includegraphics[width=\linewidth]{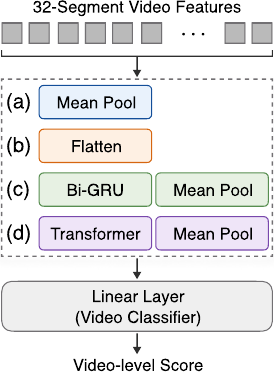}
    \caption{Architecture of the four video classifiers.}
    \vspace{-2em}
    \label{fig:probing-architecture}
\end{wrapfigure}

The maximum above uses frame-level test labels to determine the ordering
between videos.
We therefore next test whether video-level labels alone are sufficient to
produce high Micro-AUROC.
We train four video-level anomaly classifiers using the same video-level
supervision available in WSVAD.
Each classifier predicts one anomaly score for an entire video.
For Micro-AUROC evaluation, 
we repeat this video-level anomaly score across
all frames of the video.
Within-AUROC is therefore $50\%$ for every classifier.

As illustrated in~\Cref{fig:probing-architecture} the classifiers differ in how they process the video features before
predicting the video-level anomaly score.
Mean-pool averages the features over time before classification.
Sequence-linear, bi-GRU, and Transformer classifiers instead process the
standard 32-segment feature sequence~\citep{sultani2018real} before
predicting one anomaly score for the video.
We evaluate each classifier with CLIP and I3D features over three runs.

\Cref{tab:video-classifiers} shows that video-level anomaly classifiers achieve high Micro-AUROC across all four benchmarks.
The best classifier reaches $97.18\%$ on ShanghaiTech, $81.40\%$ on UCF-Crime, $90.78\%$ on XD-Violence, and $89.76\%$ on MSAD.
On ShanghaiTech with CLIP features, the video-level Transformer achieves $97.18\%$, compared with $97.74\%$ for CLIP-TSA.
Their Within-AUROC values are $50\%$ and $82.59\%$, respectively (Appendix Table~\ref{tab:public-decomp}).
Thus, similar Micro-AUROC values can accompany very different
within-video rankings even when both models use video-level
supervision and the same feature representation.

\begin{figure}[t]
\centering
\includegraphics[width=0.8\linewidth]{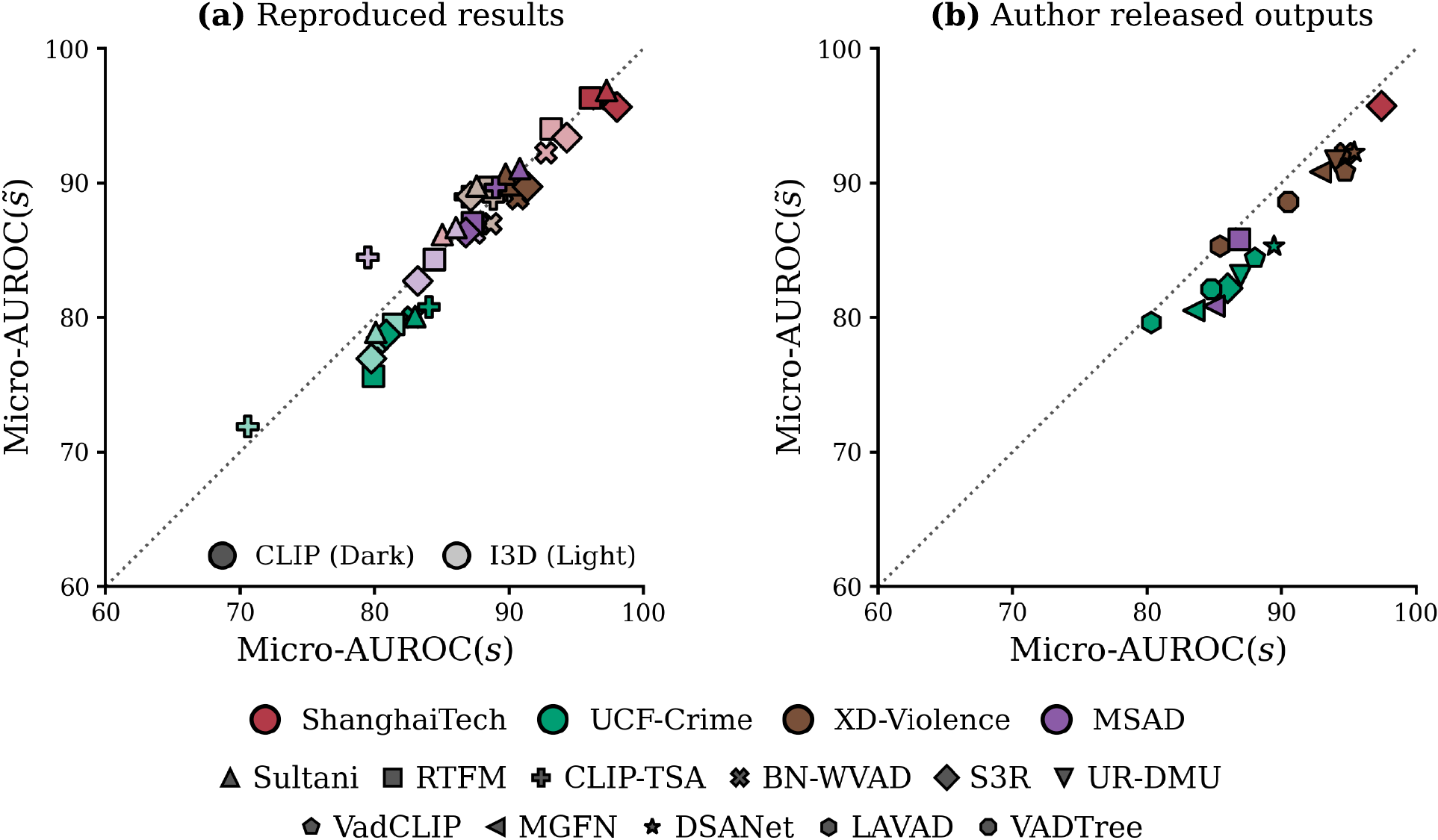}
\caption{
\textbf{Micro-AUROC remains high after video-mean replacement.}
All frame scores in each video are replaced with that video’s mean score.
The horizontal axis shows the original Micro-AUROC, and the vertical
axis shows Micro-AUROC after replacement.
(a) Each point represents one of 40 combinations of method, feature,
and benchmark, averaged over three training runs.
(b) Each point represents one of 13 released results.
The diagonal denotes no change.
}
\label{fig:video-mean-replacement}
\end{figure}

\section{Within-Video Ranking in Existing Detectors}
\label{sec:detector-evaluation}

We next examine whether high Micro-AUROC in existing detectors
reflects within-video ranking.
We evaluate the MIL ranking method of \citet{sultani2018real},
RTFM~\citep{rtfm}, S3R~\citep{s3r}, CLIP-TSA~\citep{joo2023cliptsa}, and
BN-WVAD~\citep{zhou2024batchnorm} 
with I3D and CLIP features 
on ShanghaiTech, UCF-Crime, XD-Violence, and MSAD.
Three training runs per combination give 120 runs.
We separately evaluate 13 results from author-released checkpoints
or frame-level anomaly scores.
Appendix~\ref{app:protocol} gives the experimental details.

\subsection{Video-Mean Replacement}
\label{sec:mean-replacement}

We replace all frame scores in each video with that video's mean
and evaluate the original and replaced scores on the same frames
and annotations.
Given frame-level scores $s=\{s_{it}\}_{i,t}$ and $T_i$ evaluated frames
in video $i$, \emph{video-mean replacement} gives
\begin{equation}
\tilde{s}_{it}=\bar{s}_i
=\frac{1}{T_i}\sum_{u=1}^{T_i}s_{iu}.
\label{eq:video-mean-replacement}
\end{equation}
Every within-video anomalous--normal pair is tied, so
Within-AUROC is $50\%$.
Replacement can also change cross-video comparisons, so its
effect on Micro-AUROC must be measured rather than inferred
from \Cref{eq:decomp}.

With AUROC on the $[0,1]$ scale, the retained margin above chance is
\begin{equation}
R_{\mathrm{mean}}
=\frac{\aurocmicro(\tilde{s})-0.5}{\aurocmicro(s)-0.5}.
\label{eq:video-mean-replacement-evaluation}
\end{equation}
We compute this ratio for each reproduced run or released result
before aggregation.
Values of $100\%$ indicate unchanged Micro-AUROC, and larger
values indicate an increase.

Video-mean replacement preserves a median $98.8\%$ of the
Micro-AUROC margin across 120 reproduced runs, with 116 retaining
at least $90\%$ (\Cref{fig:video-mean-replacement,tab:video-mean-replacement}).
UCF-Crime has the largest decreases, with median retention of
$92.6\%$ and a minimum of $78.1\%$.
The 13 released results retain a median $91.4\%$ and a minimum
of $88.3\%$.
High Micro-AUROC thus persists despite chance within-video ranking.
Appendix~\ref{app:detector-results} gives the complete results.

For frame-level AP, we compute retention above the anomalous-frame
fraction rather than $50\%$.
Median AP retention is $94.6\%$ on XD-Violence and $93.0\%$ on
MSAD, compared with $80.8\%$ on ShanghaiTech and $72.1\%$ on
UCF-Crime across the reproduced runs.
These are empirical results, not consequences of the AUROC
decomposition (Appendix~\ref{app:frame-ap}).

\begin{table}[t]
\centering
\caption{
Retained Micro-AUROC margin above chance after video-mean replacement.
Each reproduced run or released result contributes one
$R_{\mathrm{mean}}$ value as defined in
\Cref{eq:video-mean-replacement-evaluation}.
The final two columns count results retaining at least $85\%$
and $90\%$ of the original margin, respectively.
Reproduced runs and released results are summarized separately.
}
\label{tab:video-mean-replacement}
\small
\setlength{\tabcolsep}{4pt}
\renewcommand{\arraystretch}{1.1}
\begin{tabular*}{\linewidth}{@{\extracolsep{\fill}}lrrrrr@{}}
\toprule
Benchmark & $n$ & Median $R_{\mathrm{mean}}$
& Minimum $R_{\mathrm{mean}}$
& $R_{\mathrm{mean}}\geq85\%$
& $R_{\mathrm{mean}}\geq90\%$ \\
\midrule
\multicolumn{6}{@{}l}{\textit{Reproduced runs}} \\
ShanghaiTech & 30 & $99.3\%$  & $95.0\%$ & $30/30$   & $30/30$ \\
UCF-Crime    & 30 & $92.6\%$  & $78.1\%$ & $29/30$   & $26/30$ \\
XD-Violence  & 30 & $100.0\%$ & $94.3\%$ & $30/30$   & $30/30$ \\
MSAD         & 30 & $100.1\%$ & $95.5\%$ & $30/30$   & $30/30$ \\
All          & 120 & $98.8\%$ & $78.1\%$ & $119/120$ & $116/120$ \\
\midrule
\multicolumn{6}{@{}l}{\textit{Released checkpoints and scores}} \\
All          & 13 & $91.4\%$ & $88.3\%$ & $13/13$ & $9/13$ \\
\bottomrule
\end{tabular*}
\end{table}

\subsection{Evaluation Beyond Weak Supervision}
\label{sec:broader-evaluation}

Cross-video comparisons remain dominant beyond weak supervision,
but video-mean replacement need not preserve high Micro-AUROC.
The decomposition applies regardless of training supervision,
whereas replacement can also change cross-video score orderings.
\Cref{tab:beyond-wsvad} compares semi-supervised and training-free
detectors on the same frames before and after replacement.
For TrajVAD-P~\citep{song2026bounding} on ShanghaiTech, Micro-AUROC
is $88.59\%$ and Cross-AUROC is $88.57\%$, even though Within-AUROC
is $91.58\%$.
Only $0.814\%$ of anomalous--normal pairs come from the same video,
so cross-video comparisons dominate despite strong within-video ranking.

\begin{table}[t]
\centering
\caption{
\textbf{Cross-video comparisons remain dominant beyond weak supervision.}
We report the same detectors across multiple benchmarks.
AUROC values and the within-video pair fraction $w$ are percentages.
The final column gives Micro-AUROC after video-mean replacement,
which sets Within-AUROC to $50\%$ for every row. 
}
\label{tab:beyond-wsvad}
\small
\setlength{\tabcolsep}{3pt}
\renewcommand{\arraystretch}{1.1}
\begin{tabular*}{\linewidth}{@{\extracolsep{\fill}}llrrrrr@{}}
\toprule
\multirow{2}{*}{Benchmark}
& \multirow{2}{*}{Method}
& \multirow{2}{*}{$w$ (\%)}
& \multicolumn{3}{c}{Original scores}
& \multicolumn{1}{c}{After replacement} \\
\cmidrule(lr){4-6}
\cmidrule(l){7-7}
& & &
Micro-AUROC & Cross-AUROC & Within-AUROC & Micro-AUROC \\
\midrule
\multicolumn{7}{@{}l}{\textit{Semi-supervised}} \\
ShanghaiTech & TrajVAD-P
& 0.814 & 88.59 & 88.57 & 91.58 & 61.92 \\
ShanghaiTech & UniVAD
& 0.814 & 89.32 & 89.30 & 92.22 & 61.30 \\
Avenue & Jigsaw-VAD
& 5.079 & 90.32 & 90.21 & 92.40 & 65.27 \\
UCSD Ped2 & Jigsaw-VAD
& 6.802 & 86.41 & 85.96 & 92.57 & 61.72 \\
\midrule
\multicolumn{7}{@{}l}{\textit{Training-free}} \\
UCF-Crime & LAVAD
& 0.388 & 80.28 & 80.36 & 61.75 & 79.66 \\
UCF-Crime & VADTree
& 0.388 & 84.74 & 84.78 & 74.17 & 82.10 \\
XD-Violence & LAVAD
& 0.071 & 85.40 & 85.42 & 64.78 & 85.32 \\
XD-Violence & VADTree
& 0.071 & 90.49 & 90.50 & 80.80 & 88.62 \\
\bottomrule
\end{tabular*}
\end{table}

Replacement has little effect on LAVAD~\citep{lavad} and
VADTree~\citep{li2026vadtree} on XD-Violence.
LAVAD changes from $85.40\%$ to $85.32\%$ Micro-AUROC, whereas
TrajVAD-P on ShanghaiTech decreases from $88.59\%$ to $61.92\%$.
The test splits also permit different maximum values with one
anomaly score per video.
The optimal ordering in \Cref{prop:video-score-optimum} gives
$98.35\%$ Micro-AUROC on ShanghaiTech's weakly supervised split
but $75.78\%$ on its semi-supervised split
(\Cref{tab:vad-benchmark-composition}).
The value attained after replacement further depends on how the
detector means order the videos.

Large decreases are not confined to semi-supervised methods.
On the 44 anomaly-containing ShanghaiTech videos shared by the
weakly supervised and semi-supervised evaluations, CLIP-TSA with
CLIP features decreases from $68.58\%$ to $45.22\%$ Micro-AUROC
after replacement.
TrajVAD-P also decreases on these frames, from $89.66\%$ to $63.16\%$.
Both evaluations have $w=1.891\%$.
A small within-video pair fraction therefore does not by itself
imply high retention after replacement.
\Cref{app:broader-vad} reports the complete results and
common-test comparison.

\subsection{Detector Comparisons Across Ranking Metrics}
\label{sec:detector-comparison}

Micro-AUROC can favor a detector with worse within-video ranking.
We compare detector configurations using their original frame-level
anomaly scores under Micro-AUROC, Within-AUROC, and Macro-AUROC.
Within-AUROC weights each mixed-label video by its number of
anomalous--normal pairs, whereas Macro-AUROC gives each such video
equal weight.

On ShanghaiTech with CLIP features, S3R has higher Micro-AUROC
than CLIP-TSA, at $97.99\%$ versus $97.74\%$.
Within-AUROC instead favors CLIP-TSA, at $82.59\%$ versus $77.21\%$
for S3R.
Macro-AUROC gives the same ordering, at $81.85\%$ versus $80.02\%$.
Thus, the detector favored by pooled evaluation is worse under
both within-video metrics.

\begin{table}[t]
\centering
\caption{
\textbf{Changing the ranking metric changes detector comparisons.}
For each benchmark, 5 methods evaluated with 2 feature representations
give 10 configurations, which are ranked by their 3-run mean performance.
We compare all 45 unique pairs of configurations under Micro-AUROC
and each alternative metric.
A reversal occurs when the two metrics give opposite orderings for a pair.
Ties are excluded from the reversal count.
$\rho$ is Spearman's rank correlation.
}
\label{tab:detector-ranking-reversals}
\small
\setlength{\tabcolsep}{3pt}
\renewcommand{\arraystretch}{1.1}
\begin{tabular*}{\linewidth}{@{\extracolsep{\fill}}lrrrrrr@{}}
\toprule
& \multicolumn{2}{c}{Within-AUROC}
& \multicolumn{2}{c}{Macro-AUROC}
& \multicolumn{2}{c}{AUC$_{\mathrm{A}}$} \\
\cmidrule(lr){2-3}\cmidrule(lr){4-5}\cmidrule(l){6-7}
Benchmark & Reversals & $\rho$ & Reversals & $\rho$ & Reversals & $\rho$ \\
\midrule
ShanghaiTech & 23 & $-0.10$ & 21 & $0.10$  & 19 & $0.20$ \\
UCF-Crime    &  9 & $0.78$  & 10 & $0.68$  & 13 & $0.58$ \\
XD-Violence  & 15 & $0.44$  & 18 & $0.26$  & 25 & $-0.08$ \\
MSAD         & 31 & $-0.53$ & 28 & $-0.30$ & 27 & $-0.35$ \\
\midrule
All (180 pairs) & 78 ($43.3\%$) & -- & 77 ($42.8\%$) & -- & 84 ($46.7\%$) & -- \\
\bottomrule
\end{tabular*}
\end{table}

These reversals occur across benchmarks and feature representations.
\Cref{tab:detector-ranking-reversals} shows opposite orderings to
Micro-AUROC for 78 of 180 configuration pairs under Within-AUROC
and 77 pairs under Macro-AUROC.
On every benchmark, both within-video metrics reverse at least one
Micro-AUROC ordering among methods using CLIP features and among
methods using I3D features.
The disagreement therefore persists when the feature representation
is held fixed.

Even when the metrics agree on detector ordering, they can show very
different performance gaps.
On ShanghaiTech with CLIP features, CLIP-TSA exceeds RTFM by
$1.69$ percentage points in Micro-AUROC but by $37.60$ points in
Within-AUROC.
Micro-AUROC alone therefore establishes neither the ordering nor
the size of performance differences under within-video evaluation.
Appendix Table~\ref{tab:public-decomp} reports the configuration-level
scores.

\paragraph{Removing all-normal videos does not isolate within-video ranking.}
AUC$_{\mathrm A}$ removes all-normal videos and pools frames from
the remaining videos~\citep{lv2021localizing,lv2023unbiased}.
Using the benchmark annotations, within-video pairs still
account for only $0.24\%$--$1.90\%$ of the remaining comparisons.
Ordering videos by their ground-truth anomaly fractions gives
$77.68\%$--$85.20\%$ AUC$_{\mathrm A}$ with one score per video
(\Cref{tab:auca}).
Anomaly-only pooling changes detector rankings, but the remaining
cross-video comparisons prevent it from replacing within-video
evaluation (\Cref{tab:detector-ranking-reversals}).

\section{Implications for VAD Evaluation}
\label{sec:implications}

\paragraph{Report within-video ranking alongside pooled metrics.}
The ranking reversals in \Cref{tab:detector-ranking-reversals} show
that higher Micro-AUROC need not indicate better within-video ranking.
This distinction is rarely reflected through Macro-AUROC in the VAD papers we surveyed:
112 of 113 report frame-level AUROC or AP, whereas only 10 report
Macro-AUROC, and none of the 59 WSVAD papers do so (\Cref{tab:metric-survey}).
We therefore recommend reporting Micro-AUROC together with
Within-AUROC and Macro-AUROC to distinguish pooled from within-video ranking.
Within-AUROC weights each mixed-label video by its number of
anomalous--normal pairs, whereas Macro-AUROC gives each mixed-label
video equal weight.
Cross-AUROC and the within-video pair fraction $w$ explain how the
two pair populations contribute to Micro-AUROC under~\Cref{eq:decomp}.

\paragraph{Report the composition of the evaluated split.}
The within-video pair fraction depends on the evaluated frames and
video boundaries, not on a dataset name alone.
For example, ShanghaiTech has $w=0.140\%$ on its weakly supervised
split and $w=0.814\%$ on its semi-supervised split
(\Cref{tab:vad-benchmark-composition}).
Removing all-normal videos does not isolate within-video ranking:
after anomaly-only pooling, within-video pairs still account for only
$0.24\%$--$1.90\%$ of comparisons on the four benchmarks
(\Cref{tab:auca}).
Evaluation results should therefore identify the tested split and,
when pooled AUROC is reported, its within-video pair fraction.

\paragraph{Scope of within-video evaluation.}
Within-AUROC and Macro-AUROC measure ranking within mixed-label videos.
They do not evaluate false alarms on all-normal videos or temporal interval boundaries.
These properties require separate evaluation when they are part of the claim being made~\citep{wu2024vadclip,acharya2026road}.
Appendix~\ref{app:temporal-map} examines interval-based evaluation.

\section{Limitations}
\label{sec:limitations}

\paragraph{Statistical uncertainty in detector comparisons.}
Our comparisons of Micro-AUROC, Within-AUROC, and Macro-AUROC
report point estimates of performance differences between detectors.
We do not estimate uncertainty in these differences.
Because frames within the same video are dependent, such uncertainty
should be estimated using video-level resampling or a variance estimator
that accounts for within-video dependence~(\Cref{app:within-concentration},\citet{obuchowski1997nonparametric}).

\paragraph{Dependence on frame-level annotations.}
Multi-round re-annotation of UCF-Crime and XD-Violence documents
substantial annotator disagreement~\citep{liu2025rethinking}.
The decomposition and bound on $w$ hold for any fixed
labeling, but the numerical values of $w$ and Within-AUROC
depend on the annotated event boundaries.
Comparisons between detectors under within-video evaluation
therefore also depend on these annotations.

\section{Conclusion}
\label{sec:conclusion}

Micro-AUROC is widely used to compare VAD model performance.
However, high Micro-AUROC does not by itself establish that a detector ranks anomalous frames above normal frames within the same video.
We examine this gap by decomposing Micro-AUROC into within-video and cross-video comparisons.
Video-level classifiers and video-mean replacement then show 
that Micro-AUROC can remain high even when within-video ranking is at chance.
Evaluating the original detector scores within videos further reveals differences in within-video performance that Micro-AUROC can obscure, 
including reversals in detector rankings.
VAD evaluations should therefore distinguish pooled ranking,
within-video ranking, and interval detection, and support claims about
each with the corresponding evidence.

\newpage

\subsection*{AI Use Statement}
In this work, we used generative AI tools to implement isolated utility functions in inference code, 
and these functions were verified and tested for correctness by the authors.
Additionally, we used generative AI tools for proofreading, grammar checking, and typographical checks during writing.
We have reviewed all AI-assisted work, and every reported number was produced by the deterministic evaluation pipeline of the Reproducibility Statement.
We take responsibility for the final content of this work, including text, claims, and artifacts produced with the aid of generative AI.

\subsection*{Ethics statement}
This work analyzes existing video anomaly detection benchmarks. We do not
collect or annotate new videos, nor do we redistribute the benchmark videos.
Our analysis reports aggregate evaluation
results and does not involve person-level decisions.

\subsection*{Reproducibility statement}
Appendix~\ref{app:protocol} specifies the benchmark splits and annotations, the evaluated frame coverage,
the method--feature--benchmark training grid and seed aggregation, and the handling of author-released checkpoints and scores. 
It also describes the video-level classifier training, AUROC tie convention, and checks for video-mean replacement. 
Appendix~\ref{app:detector-results} reports the
method-level results, while Appendix~\ref{app:proofs} gives the proofs of the
AUROC decomposition and video-constant optimum.

\bibliography{iclr2027_conference}

@inproceedings{sultani2018real,
  title={Real-world anomaly detection in surveillance videos},
  author={Sultani, Waqas and Chen, Chen and Shah, Mubarak},
  booktitle={2018 IEEE/CVF Conference on Computer Vision and Pattern Recognition},
  pages={6479--6488},
  year={2018},
  organization={IEEE}
}

@inproceedings{zhou2023dual,
  title={Dual memory units with uncertainty regulation for weakly supervised video anomaly detection},
  author={Zhou, Hang and Yu, Junqing and Yang, Wei},
  booktitle={Proceedings of the AAAI conference on artificial intelligence},
  volume={37},
  number={3},
  pages={3769--3777},
  year={2023}
}

@article{zhou2024batchnorm,
  title={Batchnorm-based weakly supervised video anomaly detection},
  author={Zhou, Yixuan and Qu, Yi and Xu, Xing and Shen, Fumin and Song, Jingkuan and Shen, Heng Tao},
  journal={IEEE Transactions on Circuits and Systems for Video Technology},
  volume={34},
  number={12},
  pages={13642--13654},
  year={2024},
  publisher={IEEE}
}

@inproceedings{joo2023cliptsa,
  title={Clip-tsa: Clip-assisted temporal self-attention for weakly-supervised video anomaly detection},
  author={Joo, Hyekang Kevin and Vo, Khoa and Yamazaki, Kashu and Le, Ngan},
  booktitle={2023 IEEE International Conference on Image Processing (ICIP)},
  pages={3230--3234},
  year={2023},
  organization={IEEE}
}

@inproceedings{wu2024vadclip,
  title={Vadclip: Adapting vision-language models for weakly supervised video anomaly detection},
  author={Wu, Peng and Zhou, Xuerong and Pang, Guansong and Zhou, Lingru and Yan, Qingsen and Wang, Peng and Zhang, Yanning},
  booktitle={Proceedings of the AAAI conference on artificial intelligence},
  volume={38},
  number={6},
  pages={6074--6082},
  year={2024}
}

@article{lv2021localizing,
  title={Localizing anomalies from weakly-labeled videos},
  author={Lv, Hui and Zhou, Chuanwei and Cui, Zhen and Xu, Chunyan and Li, Yong and Yang, Jian},
  journal={IEEE transactions on image processing},
  volume={30},
  pages={4505--4515},
  year={2021},
  publisher={IEEE}
}

@inproceedings{lv2023unbiased,
  title={Unbiased multiple instance learning for weakly supervised video anomaly detection},
  author={Lv, Hui and Yue, Zhongqi and Sun, Qianru and Luo, Bin and Cui, Zhen and Zhang, Hanwang},
  booktitle={2023 IEEE/CVF Conference on Computer Vision and Pattern Recognition (CVPR)},
  pages={8022--8031},
  year={2023},
  organization={IEEE}
}

@inproceedings{acsintoae2022ubnormal,
  title={UBnormal: New Benchmark for Supervised Open-Set Video Anomaly Detection},
  author={Acsintoae, Andra and Florescu, Andrei and Georgescu, Mariana-Iuliana and Mare, Tudor and Sumedrea, Paul and Ionescu, Radu Tudor and Khan, Fahad Shahbaz and Shah, Mubarak},
  booktitle={Proceedings of the IEEE/CVF Conference on Computer Vision and Pattern Recognition},
  pages={20143--20153},
  year={2022}
}

@inproceedings{ristea2024selfdistilled,
  title={Self-Distilled Masked Auto-Encoders are Efficient Video Anomaly Detectors},
  author={Ristea, Nicolae-Catalin and Croitoru, Florinel-Alin and Ionescu, Radu Tudor and Popescu, Marius and Khan, Fahad Shahbaz and Shah, Mubarak},
  booktitle={Proceedings of the IEEE/CVF Conference on Computer Vision and Pattern Recognition},
  pages={15984--15995},
  year={2024}
}

@article{borkan2019limitations,
  title={Limitations of pinned auc for measuring unintended bias},
  author={Borkan, Daniel and Dixon, Lucas and Li, John and Sorensen, Jeffrey and Thain, Nithum and Vasserman, Lucy},
  journal={arXiv preprint arXiv:1903.02088},
  year={2019}
}

@article{kallus2019fairness,
  title={The fairness of risk scores beyond classification: Bipartite ranking and the xauc metric},
  author={Kallus, Nathan and Zhou, Angela},
  journal={Advances in neural information processing systems},
  volume={32},
  year={2019}
}

@article{obuchowski1997nonparametric,
  title={Nonparametric analysis of clustered ROC curve data},
  author={Obuchowski, Nancy A},
  journal={Biometrics},
  pages={567--578},
  year={1997},
  publisher={JSTOR}
}

@inproceedings{vogel2021learning,
  title={Learning fair scoring functions: Bipartite ranking under roc-based fairness constraints},
  author={Vogel, Robin and Bellet, Aur{\'e}lien and Cl{\'e}men{\c{c}}on, Stephan},
  booktitle={International conference on artificial intelligence and statistics},
  pages={784--792},
  year={2021},
  organization={PMLR}
}

@article{fawcett2007pav,
  title={PAV and the ROC convex hull},
  author={Fawcett, Tom and Niculescu-Mizil, Alexandru},
  journal={Machine Learning},
  volume={68},
  number={1},
  pages={97--106},
  year={2007},
  publisher={Springer}
}

@article{clemenccon2008ranking,
  title={Ranking and empirical minimization of U-statistics},
  author={Cl{\'e}men{\c{c}}on, St{\'e}phan and Lugosi, G{\'a}bor and Vayatis, Nicolas},
  year={2008}
}

@article{menon2016bipartite,
  title={Bipartite ranking: a risk-theoretic perspective},
  author={Menon, Aditya Krishna and Williamson, Robert C},
  journal={Journal of Machine Learning Research},
  volume={17},
  number={195},
  pages={1--102},
  year={2016}
}

@article{liu2025rethinking,
  title={Rethinking metrics and benchmarks of video anomaly detection},
  author={Liu, Zihao and Wu, Xiaoyu and Li, Wenna and Yang, Linlin and Wang, Shengjin},
  journal={arXiv preprint arXiv:2505.19022},
  year={2025}
}

@inproceedings{msl,
  title={Self-training multi-sequence learning with transformer for weakly supervised video anomaly detection},
  author={Li, Shuo and Liu, Fang and Jiao, Licheng},
  booktitle={Proceedings of the AAAI conference on artificial intelligence},
  volume={36},
  number={2},
  pages={1395--1403},
  year={2022}
}

@inproceedings{s3r,
  title={Self-supervised sparse representation for video anomaly detection},
  author={Wu, Jhih-Ciang and Hsieh, He-Yen and Chen, Ding-Jie and Fuh, Chiou-Shann and Liu, Tyng-Luh},
  booktitle={European Conference on Computer Vision},
  pages={729--745},
  year={2022},
  organization={Springer}
}

@inproceedings{shao2025eventvad,
  title={Eventvad: Training-free event-aware video anomaly detection},
  author={Shao, Yihua and He, Haojin and Li, Sijie and Chen, Siyu and Long, Xinwei and Zeng, Fanhu and Fan, Yuxuan and Zhang, Muyang and Yan, Ziyang and Ma, Ao and others},
  booktitle={Proceedings of the 33rd ACM International Conference on Multimedia},
  pages={2586--2595},
  year={2025}
}

@inproceedings{rtfm,
  title={Weakly-supervised video anomaly detection with robust temporal feature magnitude learning},
  author={Tian, Yu and Pang, Guansong and Chen, Yuanhong and Singh, Rajvinder and Verjans, Johan W and Carneiro, Gustavo},
  booktitle={Proceedings of the IEEE/CVF international conference on computer vision},
  pages={4975--4986},
  year={2021}
}

@inproceedings{lavad,
  title={Harnessing large language models for training-free video anomaly detection},
  author={Zanella, Luca and Menapace, Willi and Mancini, Massimiliano and Wang, Yiming and Ricci, Elisa},
  booktitle={2024 IEEE/CVF Conference on Computer Vision and Pattern Recognition (CVPR)},
  pages={18527--18536},
  year={2024},
  organization={IEEE}
}

@inproceedings{yin2026learning,
  title={Learning to tell apart: Weakly supervised video anomaly detection via disentangled semantic alignment},
  author={Yin, Wenti and Zhang, Huaxin and Wang, Xiang and Lu, Yuqing and Zhang, Yicheng and Gong, Bingquan and Zuo, Jialong and Yu, Li and Gao, Changxin and Sang, Nong},
  booktitle={Proceedings of the AAAI Conference on Artificial Intelligence},
  volume={40},
  number={14},
  pages={12027--12035},
  year={2026}
}

@inproceedings{chen2024prompt,
  title={Prompt-enhanced multiple instance learning for weakly supervised video anomaly detection},
  author={Chen, Junxi and Li, Liang and Su, Li and Zha, Zheng-jun and Huang, Qingming},
  booktitle={2024 IEEE/CVF Conference on Computer Vision and Pattern Recognition (CVPR)},
  pages={18319--18329},
  year={2024},
  organization={IEEE}
}

@article{song2026rethinking,
  title={Rethinking Open-World Video Anomaly Detection: Diagnosing Definition Blindness},
  author={Song, Inpyo and Lee, Jangwon},
  journal={arXiv preprint arXiv:2607.20780},
  year={2026}
}

@inproceedings{acharya2026road,
  title={The Road Less Seen: Segment Exploration for Weakly Supervised Video Anomaly Detection},
  author={Acharya, Anusha and Sapkota, Hitesh and Yu, Qi and Liu, Xumin},
  booktitle={Proceedings of the IEEE/CVF Conference on Computer Vision and Pattern Recognition},
  pages={14147--14156},
  year={2026}
}

@inproceedings{xu2026tlma,
  title={TLMA: Mitigating the Impact of Weakly Labeled Information for Video Anomaly Detection},
  author={Xu, Rong and Wang, Runqi and Zhang, Yingjun and Tao, Tao and Li, Xiaomeng and Jing, Liping},
  booktitle={Proceedings of the IEEE/CVF Conference on Computer Vision and Pattern Recognition},
  pages={35597--35606},
  year={2026}
}

@InProceedings{chu2026reba,
    author    = {Chu, Chengxi and Japar, Nurul and Lim, Chee Kau},
    title     = {REBA: Residual Mixture-of-Experts and Bidirectional Video-Text Alignment for Better Fine-grained Weakly Supervised Video Anomaly Detection},
    booktitle = {Proceedings of the IEEE/CVF Conference on Computer Vision and Pattern Recognition (CVPR) Findings},
    month     = {June},
    year      = {2026},
    pages     = {8280-8290}
}

@InProceedings{sun2026mome,
    author    = {Sun, Bo and Chen, Junxi and Wu, Zhe and Gao, Feng and Yang, Fan and Su, Li and Wang, Yaowei},
    title     = {Joint Learning of General and Diverse Patterns with Mixture of Memory Experts for Weakly-Supervised Video Anomaly Detection},
    booktitle = {Proceedings of the IEEE/CVF Conference on Computer Vision and Pattern Recognition (CVPR)},
    month     = {June},
    year      = {2026},
    pages     = {35638-35647}
}

@inproceedings{zhong2019graph,
  title={Graph convolutional label noise cleaner: Train a plug-and-play action classifier for anomaly detection},
  author={Zhong, Jia-Xing and Li, Nannan and Kong, Weijie and Liu, Shan and Li, Thomas H and Li, Ge},
  booktitle={2019 IEEE/CVF conference on computer vision and pattern recognition (CVPR)},
  pages={1237--1246},
  year={2019},
  organization={IEEE}
}

@inproceedings{wu2020not,
  title={Not only look, but also listen: Learning multimodal violence detection under weak supervision},
  author={Wu, Peng and Liu, Jing and Shi, Yujia and Sun, Yujia and Shao, Fangtao and Wu, Zhaoyang and Yang, Zhiwei},
  booktitle={European conference on computer vision},
  pages={322--339},
  year={2020},
  organization={Springer}
}

@article{wu2022weakly,
  title={Weakly supervised audio-visual violence detection},
  author={Wu, Peng and Liu, Xiaotao and Liu, Jing},
  journal={IEEE Transactions on Multimedia},
  volume={25},
  pages={1674--1685},
  year={2022},
  publisher={IEEE}
}

@inproceedings{yu2022modality,
  title={Modality-aware contrastive instance learning with self-distillation for weakly-supervised audio-visual violence detection},
  author={Yu, Jiashuo and Liu, Jinyu and Cheng, Ying and Feng, Rui and Zhang, Yuejie},
  booktitle={Proceedings of the 30th ACM international conference on multimedia},
  pages={6278--6287},
  year={2022}
}

@InProceedings{panariello2022consistency,
    author    = {Panariello, Aniello and Porrello, Angelo and Calderara, Simone and Cucchiara, Rita},
    title     = {Consistency-based Self-supervised Learning for Temporal Anomaly Localization},
    booktitle = {European Conference on Computer Vision (ECCV) Workshops},
    year      = {2022}
}

@inproceedings{zhang2023exploiting,
  title={Exploiting completeness and uncertainty of pseudo labels for weakly supervised video anomaly detection},
  author={Zhang, Chen and Li, Guorong and Qi, Yuankai and Wang, Shuhui and Qing, Laiyun and Huang, Qingming and Yang, Ming-Hsuan},
  booktitle={2023 IEEE/CVF Conference on Computer Vision and Pattern Recognition (CVPR)},
  pages={16271--16280},
  year={2023},
  organization={IEEE}
}

@article{pang2023audiovisual,
  title={Audiovisual dependency attention for violence detection in videos},
  author={Pang, Wenfeng and Xie, Wei and He, Qianhua and Li, Yanxiong and Yang, Jichen},
  journal={IEEE Transactions on Multimedia},
  volume={25},
  pages={4922--4932},
  year={2023},
  publisher={IEEE}
}

@inproceedings{tan2024overlooked,
  title={Overlooked video classification in weakly supervised video anomaly detection},
  author={Tan, Weijun and Yao, Qi and Liu, Jingfeng},
  booktitle={2024 IEEE/CVF Winter Conference on Applications of Computer Vision Workshops (WACVW)},
  pages={212--220},
  year={2024},
  organization={IEEE}
}

@inproceedings{ghadiya2024crossmodal,
  title={Cross-modal fusion and attention mechanism for weakly supervised video anomaly detection},
  author={Ghadiya, Ayush and Kar, Purbayan and Chudasama, Vishal and Wasnik, Pankaj},
  booktitle={2024 IEEE/CVF Conference on Computer Vision and Pattern Recognition Workshops (CVPRW)},
  pages={1965--1974},
  year={2024},
  organization={IEEE}
}

@inproceedings{wu2024weakly,
  title={Weakly supervised video anomaly detection and localization with spatio-temporal prompts},
  author={Wu, Peng and Zhou, Xuerong and Pang, Guansong and Yang, Zhiwei and Yan, Qingsen and Wang, Peng and Zhang, Yanning},
  booktitle={Proceedings of the 32nd ACM International Conference on Multimedia},
  pages={9301--9310},
  year={2024}
}

@inproceedings{jain2024crossdomain,
  title={Cross-domain learning for video anomaly detection with limited supervision},
  author={Jain, Yashika and Dabouei, Ali and Xu, Min},
  booktitle={European Conference on Computer Vision},
  pages={468--484},
  year={2024},
  organization={Springer}
}

@article{cho2024towards,
  title={Towards multi-domain learning for generalizable video anomaly detection},
  author={Cho, MyeongAh and Kim, Taeoh and Shim, Minho and Wee, Dongyoon and Lee, Sangyoun},
  journal={Advances in Neural Information Processing Systems},
  volume={37},
  pages={50256--50284},
  year={2024}
}

@article{leng2024beyond,
  title={Beyond euclidean: Dual-space representation learning for weakly supervised video violence detection},
  author={Leng, Jiaxu and Wu, Zhanjie and Tan, Mingpi and Liu, Yiran and Gan, Ji and Chen, Haosheng and Gao, Xinbo},
  journal={Advances in Neural Information Processing Systems},
  volume={37},
  pages={17373--17397},
  year={2024}
}

@Article{sun2024multimodal,
    author    = {Sun, Wenwen and Cao, Lin and Guo, Yanan and Du, Kangning},
    title     = {Multimodal and Multiscale Feature Fusion for Weakly Supervised Video Anomaly Detection},
    journal   = {Scientific Reports},
    year      = {2024}
}

@inproceedings{ye2025vera,
  title={Vera: Explainable video anomaly detection via verbalized learning of vision-language models},
  author={Ye, Muchao and Liu, Weiyang and He, Pan},
  booktitle={2025 IEEE/CVF Conference on Computer Vision and Pattern Recognition (CVPR)},
  pages={8679--8688},
  year={2025},
  organization={IEEE}
}

@article{qiu2025learning,
  title={Learning opposite prompts for weakly supervised video anomaly detection},
  author={Qiu, Helei and Hou, Biao and Cui, Yanyu},
  journal={Knowledge-Based Systems},
  volume={324},
  pages={113600},
  year={2025},
  publisher={Elsevier}
}

@article{jin2025aligning,
  title={Aligning First, Then Fusing: A novel weakly supervised multimodal violence detection method},
  author={Jin, Wenping and Zhu, Li and Sun, Jing},
  journal={Knowledge-Based Systems},
  volume={322},
  pages={113709},
  year={2025},
  publisher={Elsevier}
}

@article{li2025wsvadclip,
  title={Wsvad-clip: Temporally aware and prompt learning with clip for weakly supervised video anomaly detection},
  author={Li, Min and Sang, Jing and Lu, Yuanyao and Du, Lina},
  journal={Journal of Imaging},
  volume={11},
  number={10},
  pages={354},
  year={2025},
  publisher={MDPI}
}

@article{meng2025audiovisual,
  title={Audio-visual collaborative learning for weakly supervised video anomaly detection},
  author={Meng, Jingke and Tian, Huilin and Lin, Ge and Hu, Jian-Fang and Zheng, Wei-Shi},
  journal={IEEE Transactions on Multimedia},
  year={2025},
  publisher={IEEE}
}

@article{chu2025scene,
  title={Scene-dependent video anomaly detection: A benchmark and weakly supervised model},
  author={Chu, Chengxi and Japar, Nurul and Lim, Chee Kau},
  journal={Alexandria Engineering Journal},
  volume={133},
  pages={477--486},
  year={2025},
  publisher={Elsevier}
}

@inproceedings{damicantonio2025mixture,
  title={Mixture of experts guided by gaussian splatters matters: A new approach to weakly-supervised video anomaly detection},
  author={Amicantonio, Giacomo D' and Majhi, Snehashis and Kong, Quan and Garattoni, Lorenzo and Francesca, Gianpiero and Br{\'e}mond, Fran{\c{c}}ois and Bondarev, Egor},
  booktitle={Proceedings of the IEEE/CVF International Conference on Computer Vision},
  pages={10275--10285},
  year={2025}
}

@inproceedings{zhao2026learning,
  title={Learning from Noisy Supervision: A Denoising-Debiasing Framework for Weakly Supervised Video Anomaly Detection},
  author={Zhao, Yaxin and Wang, Yang and Guo, Wenya and Xu, Sihan and Cai, Xiangrui and Lin, Xi and Zhang, Ying and Yuan, Xiaojie},
  booktitle={Proceedings of the IEEE/CVF Conference on Computer Vision and Pattern Recognition},
  pages={21326--21335},
  year={2026}
}

@article{sun2026enhancing,
  title={Enhancing Weakly Supervised Multimodal Video Anomaly Detection through Text Guidance},
  author={Sun, Shengyang and Hua, Jiashen and Feng, Junyi and Gong, Xiaojin},
  journal={IEEE Transactions on Multimedia},
  year={2026},
  publisher={IEEE}
}

@inproceedings{liu2018future,
  title={Future frame prediction for anomaly detection--a new baseline},
  author={Liu, Wen and Luo, Weixin and Lian, Dongze and Gao, Shenghua},
  booktitle={Proceedings of the IEEE conference on computer vision and pattern recognition},
  pages={6536--6545},
  year={2018}
}

@article{song2026vlm,
  title={A VLM Answer Is Not an Anomaly Score: Rank Compression in Training-Free Video Anomaly Detection},
  author={Song, Inpyo and Lee, Jangwon},
  journal={arXiv preprint arXiv:2608.21244},
  year={2026}
}

@inproceedings{carreira2017quo,
  title={Quo vadis, action recognition? a new model and the kinetics dataset},
  author={Carreira, Joao and Zisserman, Andrew},
  booktitle={2017 IEEE conference on computer vision and pattern recognition (CVPR)},
  pages={4724--4733},
  year={2017},
  organization={IEEE}
}

@inproceedings{wang2022video,
  title={Video anomaly detection by solving decoupled spatio-temporal jigsaw puzzles},
  author={Wang, Guodong and Wang, Yunhong and Qin, Jie and Zhang, Dongming and Bao, Xiuguo and Huang, Di},
  booktitle={European Conference on Computer Vision},
  pages={494--511},
  year={2022},
  organization={Springer}
}

@inproceedings{msad2024,
title = {Advancing Video Anomaly Detection: A Concise Review and a New Dataset},
author = {Liyun Zhu and Lei Wang and Arjun Raj and Tom Gedeon and Chen Chen},
booktitle = {The Thirty-eighth Conference on Neural Information Processing Systems Datasets and Benchmarks Track},
year = {2024}
}

@article{li2013anomaly,
  title     = {Anomaly detection and localization in crowded scenes},
  author    = {Li, Weixin and Mahadevan, Vijay and Vasconcelos, Nuno},
  journal   = {IEEE Transactions on Pattern Analysis and Machine Intelligence},
  year      = {2013},
  publisher = {IEEE}
}

@inproceedings{lu2013abnormal,
  title     = {Abnormal event detection at 150 fps in matlab},
  author    = {Lu, Cewu and Shi, Jianping and Jia, Jiaya},
  booktitle = {Proceedings of the IEEE/CVF International Conference on Computer Vision},
  year      = {2013}
}

@misc{mgfn,
  title={MGFN: Magnitude-Contrastive Glance-and-Focus Network for Weakly-Supervised Video Anomaly Detection}, 
  author={Yingxian Chen and Zhengzhe Liu and Baoheng Zhang and Wilton Fok and Xiaojuan Qi and Yik-Chung Wu},
  year={2022},
  eprint={2211.15098},
  archivePrefix={arXiv},
  primaryClass={cs.CV}
}

@inproceedings{karim2024realtime,
  title={Real-time weakly supervised video anomaly detection},
  author={Karim, Hamza and Doshi, Keval and Yilmaz, Yasin},
  booktitle={2024 IEEE/CVF Winter Conference on Applications of Computer Vision (WACV)},
  pages={6834--6842},
  year={2024},
  organization={IEEE}
}

@inproceedings{cao2023benchmark,
  title={A new comprehensive benchmark for semi-supervised video anomaly detection and anticipation},
  author={Cao, Congqi and Lu, Yue and Wang, Peng and Zhang, Yanning},
  booktitle={2023 IEEE/CVF Conference on Computer Vision and Pattern Recognition (CVPR)},
  pages={20392--20401},
  year={2023},
  organization={IEEE}
}

@inproceedings{zhang2024multiscale,
  title={Multi-scale video anomaly detection by multi-grained spatio-temporal representation learning},
  author={Zhang, Menghao and Wang, Jingyu and Qi, Qi and Sun, Haifeng and Zhuang, Zirui and Ren, Pengfei and Ma, Ruilong and Liao, Jianxin},
  booktitle={2024 IEEE/CVF Conference on Computer Vision and Pattern Recognition (CVPR)},
  pages={17385--17394},
  year={2024},
  organization={IEEE}
}

@inproceedings{micorek2024mulde,
  title={Mulde: Multiscale log-density estimation via denoising score matching for video anomaly detection},
  author={Micorek, Jakub and Possegger, Horst and Narnhofer, Dominik and Bischof, Horst and Kozinski, Mateusz},
  booktitle={Proceedings of the IEEE/CVF conference on computer vision and pattern recognition},
  pages={18868--18877},
  year={2024}
}

@inproceedings{jiang2025local,
  title={Local patterns generalize better for novel anomalies},
  author={Jiang, Yalong},
  booktitle={International Conference on Learning Representations},
  volume={2025},
  pages={49969--49995},
  year={2025}
}

@inproceedings{zhang2025autoregressive,
  title={Autoregressive denoising score matching is a good video anomaly detector},
  author={Zhang, Hanwen and Cao, Congqi and Lv, Qinyi and Min, Lingtong and Zhang, Yanning},
  booktitle={2025 IEEE/CVF International Conference on Computer Vision (ICCV)},
  pages={12057--12067},
  year={2025},
  organization={IEEE}
}

@inproceedings{park2020memory,
  title={Learning memory-guided normality for anomaly detection},
  author={Park, Hyunjong and Noh, Jongyoun and Ham, Bumsub},
  booktitle={2020 IEEE/CVF Conference on Computer Vision and Pattern Recognition (CVPR)},
  pages={14360--14369},
  year={2020},
  organization={IEEE}
}

@article{barbalau2023ssmtl,
  title={SSMTL++: Revisiting self-supervised multi-task learning for video anomaly detection},
  author={Barbalau, Antonio and Ionescu, Radu Tudor and Georgescu, Mariana-Iuliana and Dueholm, Jacob and Ramachandra, Bharathkumar and Nasrollahi, Kamal and Khan, Fahad Shahbaz and Moeslund, Thomas B and Shah, Mubarak},
  journal={Computer Vision and Image Understanding},
  volume={229},
  pages={103656},
  year={2023},
  publisher={Elsevier}
}

@inproceedings{yang2024followrules,
  title={Follow the rules: reasoning for video anomaly detection with large language models},
  author={Yang, Yuchen and Lee, Kwonjoon and Dariush, Behzad and Cao, Yinzhi and Lo, Shao-Yuan},
  booktitle={European Conference on Computer Vision},
  pages={304--322},
  year={2024},
  organization={Springer}
}

@inproceedings{ahn2024videopatchcore,
  title={Videopatchcore: An effective method to memorize normality for video anomaly detection},
  author={Ahn, Sunghyun and Jo, Youngwan and Lee, Kijung and Park, Sanghyun},
  booktitle={Asian Conference on Computer Vision},
  pages={312--328},
  year={2024},
  organization={Springer}
}

@inproceedings{ahn2026anyanomaly,
  title={Anyanomaly: Zero-shot customizable video anomaly detection with lvlm},
  author={Ahn, Sunghyun and Jo, Youngwan and Lee, Kijung and Kwon, Sein and Hong, Inpyo and Park, Sanghyun},
  booktitle={2026 IEEE/CVF Winter Conference on Applications of Computer Vision (WACV)},
  pages={3026--3035},
  year={2026},
  organization={IEEE}
}

@article{li2026vadtree,
  title={Vadtree: Explainable training-free video anomaly detection via hierarchical granularity-aware tree},
  author={Li, Wenlong and Xu, Yifei and Rao, Yuan and Wang, Zhenhua and Deng, Shuiguang},
  journal={Advances in Neural Information Processing Systems},
  volume={38},
  pages={148372--148404},
  year={2026}
}

@inproceedings{feng2021mist,
  title={Mist: Multiple instance self-training framework for video anomaly detection},
  author={Feng, Jia-Chang and Hong, Fa-Ting and Zheng, Wei-Shi},
  booktitle={2021 IEEE/CVF Conference on Computer Vision and Pattern Recognition (CVPR)},
  pages={14004--14013},
  year={2021},
  organization={IEEE}
}

@inproceedings{dalvi2025distilling,
  title={Distilling aggregated knowledge for weakly-supervised video anomaly detection},
  author={Dalvi, Jash and Dabouei, Ali and Dhanuka, Gunjan and Xu, Min},
  booktitle={2025 IEEE/CVF Winter Conference on Applications of Computer Vision (WACV)},
  pages={5439--5448},
  year={2025},
  organization={IEEE}
}

@inproceedings{xu2025suppression,
  title={Discriminative score suppression for weakly supervised video anomaly detection},
  author={Xu, Chen and Li, Chunguo and Xing, Hongjie},
  booktitle={2025 IEEE/CVF Winter Conference on Applications of Computer Vision (WACV)},
  pages={9587--9596},
  year={2025},
  organization={IEEE}
}

@article{wang2026las,
  title={Weakly supervised video anomaly detection with anomaly-connected components and intention reasoning},
  author={Wang, Yu and Zhao, Shengjie},
  journal={arXiv preprint arXiv:2603.00550},
  year={2026}
}

@inproceedings{yang2024normality,
  title={Text prompt with normality guidance for weakly supervised video anomaly detection},
  author={Yang, Zhiwei and Liu, Jing and Wu, Peng},
  booktitle={2024 IEEE/CVF Conference on Computer Vision and Pattern Recognition (CVPR)},
  pages={18899--18908},
  year={2024},
  organization={IEEE}
}

@inproceedings{wu2024openvocab,
  title={Open-vocabulary video anomaly detection},
  author={Wu, Peng and Zhou, Xuerong and Pang, Guansong and Sun, Yujia and Liu, Jing and Wang, Peng and Zhang, Yanning},
  booktitle={2024 IEEE/CVF Conference on Computer Vision and Pattern Recognition (CVPR)},
  pages={18297--18307},
  year={2024},
  organization={IEEE}
}

@inproceedings{zheng2026logical,
  title={Logical Consistency Optimization for Few-Shot Weakly Supervised Video Anomaly Detection},
  author={Zheng, Hantao and Han, Ning and Zeng, Yawen and Zhu, Hegui and Chen, Hao},
  booktitle={Proceedings of the IEEE/CVF Conference on Computer Vision and Pattern Recognition},
  pages={9045--9054},
  year={2026}
}

@inproceedings{hashimoto2026cade,
  title={CADE: Continual weakly-supervised video anomaly detection with ensembles},
  author={Hashimoto, Satoshi and Konishi, Tatsuya and Kaichi, Tomoya and Matsumoto, Kazunori and Kurokawa, Mori},
  booktitle={2026 IEEE/CVF Winter Conference on Applications of Computer Vision (WACV)},
  pages={708--717},
  year={2026},
  organization={IEEE}
}

@inproceedings{morais2019regularity,
  title={Learning regularity in skeleton trajectories for anomaly detection in videos},
  author={Morais, Romero and Le, Vuong and Tran, Truyen and Saha, Budhaditya and Mansour, Moussa and Venkatesh, Svetha},
  booktitle={2019 IEEE/CVF Conference on Computer Vision and Pattern Recognition (CVPR)},
  pages={11988--11996},
  year={2019},
  organization={IEEE}
}

@inproceedings{ristea2022sspcab,
  title={Self-supervised predictive convolutional attentive block for anomaly detection},
  author={Ristea, Nicolae-C{\u{a}}t{\u{a}}lin and Madan, Neelu and Ionescu, Radu Tudor and Nasrollahi, Kamal and Khan, Fahad Shahbaz and Moeslund, Thomas B and Shah, Mubarak},
  booktitle={2022 IEEE/CVF Conference on Computer Vision and Pattern Recognition (CVPR)},
  pages={13566--13576},
  year={2022},
  organization={IEEE}
}

@inproceedings{gong2019memorizing,
  title={Memorizing normality to detect anomaly: Memory-augmented deep autoencoder for unsupervised anomaly detection},
  author={Gong, Dong and Liu, Lingqiao and Le, Vuong and Saha, Budhaditya and Mansour, Moussa Reda and Venkatesh, Svetha and Hengel, Anton van den},
  booktitle={Proceedings of the IEEE/CVF international conference on computer vision},
  pages={1705--1714},
  year={2019}
}

@inproceedings{ionescu2019object,
  title={Object-centric auto-encoders and dummy anomalies for abnormal event detection in video},
  author={Ionescu, Radu Tudor and Khan, Fahad Shahbaz and Georgescu, Mariana-Iuliana and Shao, Ling},
  booktitle={2019 IEEE/CVF conference on computer vision and pattern recognition (CVPR)},
  pages={7834--7843},
  year={2019},
  organization={IEEE}
}

@inproceedings{nguyen2019appearance,
  title={Anomaly detection in video sequence with appearance-motion correspondence},
  author={Nguyen, Trong-Nguyen and Meunier, Jean},
  booktitle={Proceedings of the IEEE/CVF international conference on computer vision},
  pages={1273--1283},
  year={2019}
}

@inproceedings{georgescu2021multitask,
  title={Anomaly detection in video via self-supervised and multi-task learning},
  author={Georgescu, Mariana-Iuliana and Barbalau, Antonio and Ionescu, Radu Tudor and Khan, Fahad Shahbaz and Popescu, Marius and Shah, Mubarak},
  booktitle={Proceedings of the IEEE/CVF conference on computer vision and pattern recognition},
  pages={12742--12752},
  year={2021}
}

@inproceedings{astrid2021synthetic,
  title={Synthetic temporal anomaly guided end-to-end video anomaly detection},
  author={Astrid, Marcella and Zaheer, Muhammad Zaigham and Lee, Seung-Ik},
  booktitle={2021 IEEE/CVF International Conference on Computer Vision Workshops (ICCVW)},
  pages={207--214},
  year={2021},
  organization={IEEE}
}

@inproceedings{park2022fastano,
  title={FastAno: Fast anomaly detection via spatio-temporal patch transformation},
  author={Park, Chaewon and Cho, MyeongAh and Lee, Minhyeok and Lee, Sangyoun},
  booktitle={2022 IEEE/CVF Winter Conference on Applications of Computer Vision (WACV)},
  pages={1908--1918},
  year={2022},
  organization={IEEE}
}

@inproceedings{szymanowicz2021xman,
  title={X-MAN: Explaining multiple sources of anomalies in video},
  author={Szymanowicz, Stanislaw and Charles, James and Cipolla, Roberto},
  booktitle={2021 IEEE/CVF Conference on Computer Vision and Pattern Recognition Workshops (CVPRW)},
  pages={3218--3226},
  year={2021},
  organization={IEEE}
}

@inproceedings{yang2023restoration,
  title={Video event restoration based on keyframes for video anomaly detection},
  author={Yang, Zhiwei and Liu, Jing and Wu, Zhaoyang and Wu, Peng and Liu, Xiaotao},
  booktitle={2023 IEEE/CVF Conference on Computer Vision and Pattern Recognition (CVPR)},
  pages={14592--14601},
  year={2023},
  organization={IEEE}
}

@inproceedings{yan2023diffusion,
  title={Feature prediction diffusion model for video anomaly detection},
  author={Yan, Cheng and Zhang, Shiyu and Liu, Yang and Pang, Guansong and Wang, Wenjun},
  booktitle={2023 IEEE/CVF International Conference on Computer Vision (ICCV)},
  pages={5504--5514},
  year={2023},
  organization={IEEE}
}

@inproceedings{shi2023pretext,
  title={Video anomaly detection via sequentially learning multiple pretext tasks},
  author={Shi, Chenrui and Sun, Che and Wu, Yuwei and Jia, Yunde},
  booktitle={2023 IEEE/CVF International Conference on Computer Vision (ICCV)},
  pages={10296--10306},
  year={2023},
  organization={IEEE}
}

@inproceedings{baradaran2023attention,
  title={Multi-task learning based video anomaly detection with attention},
  author={Baradaran, Mohammad and Bergevin, Robert},
  booktitle={2023 IEEE/CVF Conference on Computer Vision and Pattern Recognition Workshops (CVPRW)},
  pages={2886--2896},
  year={2023},
  organization={IEEE}
}

@inproceedings{lappas2024distinction,
  title={Dynamic distinction learning: adaptive pseudo anomalies for video anomaly detection},
  author={Lappas, Demetris and Argyriou, Vasileios and Makris, Dimitrios},
  booktitle={2024 IEEE/CVF Conference on Computer Vision and Pattern Recognition Workshops (CVPRW)},
  pages={3961--3970},
  year={2024},
  organization={IEEE}
}

@inproceedings{rai2024pseudo,
  title={Video anomaly detection via spatio-temporal pseudo-anomaly generation: A unified approach},
  author={Rai, Ayush K and Krishna, Tarun and Hu, Feiyan and Drimbarean, Alexandru and McGuinness, Kevin and Smeaton, Alan F and O'connor, Noel E},
  booktitle={Proceedings of the IEEE/CVF conference on computer vision and pattern recognition},
  pages={3887--3899},
  year={2024}
}

@inproceedings{hu2025noiseresistant,
  title={Noise-resistant video anomaly detection via RGB error-guided multiscale predictive coding and dynamic memory},
  author={Hu, Han and Du, Wenli and Liao, Peng and Wang, Bing and Fan, Siyuan},
  booktitle={2025 IEEE/CVF Conference on Computer Vision and Pattern Recognition (CVPR)},
  pages={19109--19119},
  year={2025},
  organization={IEEE}
}

@article{sun2024eventdriven,
  title={Event-driven weakly supervised video anomaly detection},
  author={Sun, Shengyang and Gong, Xiaojin},
  journal={Image and Vision Computing},
  volume={149},
  pages={105169},
  year={2024},
  publisher={Elsevier}
}

@article{lee2024cognitive,
  title={Cognitive Refined Augmentation for Video Anomaly Detection in Weak Supervision},
  author={Lee, Junyeop and Koo, Hyunbon and Kim, Seongjun and Ko, Hanseok},
  journal={Sensors},
  volume={24},
  number={1},
  pages={58},
  year={2023},
  publisher={MDPI}
}

@article{qiu2025temporal,
  title={Temporal context and representative feature learning for weakly supervised video anomaly detection},
  author={Qiu, Helei and Hou, Biao},
  journal={Journal of Information and Intelligence},
  year={2025},
  publisher={Elsevier}
}

@article{he2025spmil,
  title={A self-paced multiple instance learning framework for weakly supervised video anomaly detection},
  author={He, Ping and Li, Huibin and Han, Miaolin},
  journal={Applied Sciences},
  volume={15},
  number={3},
  pages={1049},
  year={2025},
  publisher={MDPI}
}

@article{yang2026eks,
  title={Explicit knowledge-structured weakly supervised video anomaly detection},
  author={Yang, Chen and Tang, Xue-song and Hao, Kuangrong and Li, Yubing and Wang, Zhiqi},
  journal={Digital Signal Processing},
  pages={106088},
  year={2026},
  publisher={Elsevier}
}

@article{pu2024prompt,
  title={Learning prompt-enhanced context features for weakly-supervised video anomaly detection},
  author={Pu, Yujiang and Wu, Xiaoyu and Yang, Lulu and Wang, Shengjin},
  journal={IEEE Transactions on Image Processing},
  volume={33},
  pages={4923--4936},
  year={2024},
  publisher={IEEE}
}

@article{senapati2026poinclip,
  title={PoinCLIP-VAD: a hyperbolic cross-modal fusion framework for video anomaly detection},
  author={Senapati, Debi Prasad and Pani, Santosh Kumar and Baliarsingh, Santos Kumar and Dev, Prabhu Prasad and Somula, Ramasubbareddy},
  journal={Scientific Reports},
  volume={16},
  number={1},
  pages={20341},
  year={2026},
  publisher={Nature Publishing Group}
}

@inproceedings{yu2020cloze,
  title={Cloze test helps: Effective video anomaly detection via learning to complete video events},
  author={Yu, Guang and Wang, Siqi and Cai, Zhiping and Zhu, En and Xu, Chuanfu and Yin, Jianping and Kloft, Marius},
  booktitle={Proceedings of the 28th ACM international conference on multimedia},
  pages={583--591},
  year={2020}
}

@article{cho2022flows,
  title={Unsupervised video anomaly detection via normalizing flows with implicit latent features},
  author={Cho, MyeongAh and Kim, Taeoh and Kim, Woo Jin and Cho, Suhwan and Lee, Sangyoun},
  journal={Pattern Recognition},
  volume={129},
  pages={108703},
  year={2022},
  publisher={Elsevier}
}

@inproceedings{lu2020fewshot,
  title={Few-shot scene-adaptive anomaly detection},
  author={Lu, Yiwei and Yu, Frank and Reddy, Mahesh Kumar Krishna and Wang, Yang},
  booktitle={European conference on computer vision},
  pages={125--141},
  year={2020},
  organization={Springer}
}

@article{luo2021normalgraph,
  title={Normal graph: Spatial temporal graph convolutional networks based prediction network for skeleton based video anomaly detection},
  author={Luo, Weixin and Liu, Wen and Gao, Shenghua},
  journal={Neurocomputing},
  volume={444},
  pages={332--337},
  year={2021},
  publisher={Elsevier}
}

@article{wu2022explainable,
  title={An explainable and efficient deep learning framework for video anomaly detection},
  author={Wu, Chongke and Shao, Sicong and Tunc, Cihan and Satam, Pratik and Hariri, Salim},
  journal={Cluster computing},
  volume={25},
  number={4},
  pages={2715--2737},
  year={2022},
  publisher={Springer}
}

@article{astrid2023pseudobound,
  title={Pseudobound: Limiting the anomaly reconstruction capability of one-class classifiers using pseudo anomalies},
  author={Astrid, Marcella and Zaheer, Muhammad Zaigham and Lee, Seung-Ik},
  journal={Neurocomputing},
  volume={534},
  pages={147--160},
  year={2023},
  publisher={Elsevier}
}

@article{zeng2024progressive,
  title={Progressive prediction: Video anomaly detection via multi-grained prediction},
  author={Zeng, Xianlin and Jiang, Yalong and Wang, Yufeng and Fu, Qiang and Ding, Wenrui},
  journal={IET Image Processing},
  volume={18},
  number={10},
  pages={2568--2583},
  year={2024},
  publisher={Wiley Online Library}
}

@article{wang2023ensemble,
  title={Ensemble anomaly score for video anomaly detection using denoise diffusion model and motion filters},
  author={Wang, Zhiqiang and Gu, Xiaojing and Hu, Jingyu and Gu, Xingsheng},
  journal={Neurocomputing},
  volume={553},
  pages={126589},
  year={2023},
  publisher={Elsevier}
}

@article{shao2024consistency,
  title={Consistency-constrained unsupervised video anomaly detection framework based on Co-teaching},
  author={Rajapaksha, Praboda and Crespi, Noel and Zhao, Xuechen and Wang, Mengzhu and Yin, Nan and Liu, Xinwang and Luo, Zhigang and others},
  journal={Neurocomputing},
  volume={610},
  pages={128589},
  year={2024},
  publisher={Elsevier}
}

@article{liu2024vadiffusion,
  title={VADiffusion: Compressed domain information guided conditional diffusion for video anomaly detection},
  author={Liu, Hao and He, Lijun and Zhang, Miao and Li, Fan},
  journal={IEEE Transactions on Circuits and Systems for Video Technology},
  volume={34},
  number={9},
  pages={8398--8411},
  year={2024},
  publisher={IEEE}
}

@inproceedings{zhou2025mapdm,
  title={Video anomaly detection with motion and appearance guided patch diffusion model},
  author={Zhou, Hang and Cai, Jiale and Ye, Yuteng and Feng, Yonghui and Gao, Chenxing and Yu, Junqing and Song, Zikai and Yang, Wei},
  booktitle={Proceedings of the AAAI Conference on Artificial Intelligence},
  volume={39},
  number={10},
  pages={10761--10769},
  year={2025}
}

@article{yu2026stepdiffusion,
  title={Effective video anomaly detection by step-constrained diffusion model},
  author={Yu, Guang and Xi, Junhua and Wang, Siqi and Cai, Zhiping and Deng, Kefeng and Ren, Kaijun},
  journal={Pattern Recognition},
  pages={114012},
  year={2026},
  publisher={Elsevier}
}

@article{li2026causal,
  title={Robust video anomaly detection via causal feature-guided data augmentation},
  author={Li, Zhaoyi and Shi, Chenrui and Sun, Che and Wu, Yuwei},
  journal={Computer Vision and Image Understanding},
  pages={104671},
  year={2026},
  publisher={Elsevier}
}

@inproceedings{mo2026reci,
  title={Retrieval-Guided Contextual Inference for Training-Free Video Anomaly Detection in Low-Light Scenarios},
  author={Mo, Mengjingcheng and Zheng, Jiankang and Leng, Jiaxu and Gao, Xinbo},
  booktitle={Proceedings of the 2026 International Conference on Multimedia Retrieval},
  pages={930--939},
  year={2026}
}

@article{feng2026monitor,
  title={MoniTor: Exploiting large language models with instruction for online video anomaly detection},
  author={Feng, Yue and Liu, Yingshi and Zhang, Jingrou and Qin, Jie and others},
  journal={Advances in Neural Information Processing Systems},
  volume={38},
  pages={35851--35871},
  year={2026}
}

@article{lin2025holistic,
  title={A unified reasoning framework for holistic zero-shot video anomaly analysis},
  author={Lin, Dongheng and Qu, Mengxue and Han, Kunyang and Jiao, Jianbo and Jin, Xiaojie and Wei, Yunchao},
  journal={Advances in Neural Information Processing Systems},
  volume={38},
  pages={29713--29744},
  year={2026}
}

@inproceedings{zaheer2020claws,
  title={Claws: Clustering assisted weakly supervised learning with normalcy suppression for anomalous event detection},
  author={Zaheer, Muhammad Zaigham and Mahmood, Arif and Astrid, Marcella and Lee, Seung-Ik},
  booktitle={European Conference on Computer Vision},
  pages={358--376},
  year={2020},
  organization={Springer}
}

@article{astrid2021learning,
  title={Learning not to reconstruct anomalies},
  author={Astrid, Marcella and Zaheer, Muhammad Zaigham and Lee, Jae-Yeong and Lee, Seung-Ik},
  journal={arXiv preprint arXiv:2110.09742},
  year={2021}
}

@inproceedings{cheng2024dhvad,
  title={Denoising diffusion-augmented hybrid video anomaly detection via reconstructing noised frames},
  author={Cheng, Kai and Pan, Yaning and Liu, Yang and Zeng, Xinhua and Feng, Rui},
  booktitle={Proceedings of the Thirty-Third International Joint Conference on Artificial Intelligence},
  pages={695--703},
  year={2024}
}

@article{li2024vlavad,
  title={Vision-language models assisted unsupervised video anomaly detection},
  author={Jiang, Yalong and Mao, Liquan},
  journal={arXiv preprint arXiv:2409.14109},
  year={2024}
}

@article{reiss2025attribute,
  title={An attribute-based method for video anomaly detection},
  author={Reiss, Tal and Hoshen, Yedid},
  journal={arXiv preprint arXiv:2212.00789},
  year={2022}
}

@article{mumcu2026mllm,
  title={Leveraging multimodal llm descriptions of activity for explainable semi-supervised video anomaly detection},
  author={Mumcu, Furkan and Jones, Michael J and Cherian, Anoop and Yilmaz, Yasin},
  journal={arXiv preprint arXiv:2510.14896},
  year={2025}
}

@inproceedings{zhang2026fine,
  title={Fine-VAD: Towards Fine-Grained Video Anomaly Detection via Progressive Cross-Granularity Learning},
  author={Zhang, Menghao and Zhu, Yiyan and Ren, Pengfei and Sun, Haifeng and Qi, Qi and Zhuang, Zirui and Wang, Huazheng and Zhang, Lei and Liao, Jianxin and Wang, Jingyu},
  booktitle={Proceedings of the IEEE/CVF Conference on Computer Vision and Pattern Recognition},
  pages={35514--35523},
  year={2026}
}

@inproceedings{song2025anomaly,
  title={Anomaly detection for people with visual impairments using an egocentric 360-degree camera},
  author={Song, Inpyo and Lee, Sanghyeon and Joo, Minjun and Lee, Jangwon},
  booktitle={2025 IEEE/CVF Winter Conference on Applications of Computer Vision (WACV)},
  pages={2828--2837},
  year={2025},
  organization={IEEE}
}

@article{song2026bounding,
  title={Bounding-Box Trajectories Matter for Video Anomaly Detection},
  author={Song, Inpyo and Lee, Jangwon},
  journal={arXiv preprint arXiv:2605.21957},
  year={2026}
}

@inproceedings{radford2021learning,
  title={Learning transferable visual models from natural language supervision},
  author={Radford, Alec and Kim, Jong Wook and Hallacy, Chris and Ramesh, Aditya and Goh, Gabriel and Agarwal, Sandhini and Sastry, Girish and Askell, Amanda and Mishkin, Pamela and Clark, Jack and others},
  booktitle={International conference on machine learning},
  pages={8748--8763},
  year={2021},
  organization={PmLR}
}
\bibliographystyle{iclr2027_conference}

\appendix

\section{Proofs for AUROC attribution}
\label{app:proofs}

\paragraph{Tie-aware pair contribution.}
For scores $x$ and $y$, define
\[
\tau(x,y)=\mathbf{1}[x>y]+\tfrac{1}{2}\mathbf{1}[x=y].
\]
Let $P_i$ and $N_i$ be the positive- and negative-frame index sets of video $i$. Micro-AUROC is
\[
\aurocmicro
=\frac{1}{AN}\sum_i\sum_{t\in P_i}\sum_j\sum_{u\in N_j}\tau(s_{it},s_{ju}).
\]
Let $\mathcal{M}=\{i:a_i>0\ \text{and}\ n_i>0\}$ denote the mixed-label videos. Videos
outside $\mathcal{M}$ have no same-video positive--negative pairs, but their frames remain in the
cross-video terms above.

\paragraph{Proof of Proposition~\ref{prop:decomp}.}
Partition the outer sum into terms with $i=j$ and terms with $i\neq j$. Define
\[
S_{\mathrm{within}}=\sum_{i\in\mathcal{M}}\sum_{t\in P_i}\sum_{u\in N_i}\tau(s_{it},s_{iu}),
\qquad
S_{\mathrm{cross}}=\sum_{i\neq j}\sum_{t\in P_i}\sum_{u\in N_j}\tau(s_{it},s_{ju}).
\]
Under the conditions of Proposition~\ref{prop:decomp}, the corresponding pair counts are
$W=\sum_{i\in\mathcal{M}}a_i n_i>0$ and $AN-W>0$. Therefore
\[
\aurocwithin=\frac{S_{\mathrm{within}}}{W},
\qquad
\auroccross=\frac{S_{\mathrm{cross}}}{AN-W},
\]
and
\[
\aurocmicro
=\frac{S_{\mathrm{within}}+S_{\mathrm{cross}}}{AN}
=\frac{W}{AN}\aurocwithin+\frac{AN-W}{AN}\auroccross.
\]
This is Equation~\ref{eq:decomp} with $w=W/(AN)$.
At the excluded boundaries, one normalized component is undefined. If $W=0$, Micro-AUROC reduces
to Cross-AUROC. If $AN-W=0$, it reduces to Within-AUROC.

\paragraph{Proof of Equation~\ref{eq:dilution} and Corollary~\ref{cor:maximum}.}
Since $p^+$ and $p^-$ are probability vectors,
\[
w=\sum_i p_i^+p_i^-
\leq \lVert p^+\rVert_\infty\sum_i p_i^-
=\lVert p^+\rVert_\infty.
\]
Exchanging the two classes gives $w\leq\lVert p^-\rVert_\infty$, which proves
Equation~\ref{eq:dilution}. If no video captures more than $C_+/V$ of the positive frames or more
than $C_-/V$ of the negative frames, then
$w\leq\min(C_+,C_-)/V$.

For the exact replication statement, make $k$ disjoint copies of every video. Each copy contributes
$a_i n_i$ same-video pairs, so the numerator of $w$ becomes $k\sum_i a_i n_i$. The total positive
and negative frame counts each grow by $k$, so the denominator becomes $k^2AN$. Hence the new weight
is $w/k$.

Finally, with Cross-AUROC fixed, changing $\aurocwithin$ from $0.5$ to $1$ changes Micro-AUROC
by $w(1-0.5)=w/2$. This proves Corollary~\ref{cor:maximum}.

\paragraph{Proof of Proposition~\ref{prop:video-score-optimum}.}
Assigning one anomaly score to every frame in a video ties all within-video
anomalous--normal pairs, so $\aurocwithin=0.5$. Consider two distinct videos $i$ and $j$. If $c_i>c_j$, the ordering correctly
ranks the $a_i n_j$ pairs whose positive frame is in $i$ and negative frame is in $j$. Reversing the
order instead correctly ranks $a_j n_i$ pairs. Therefore placing $i$ above $j$ is weakly preferable
exactly when
\[
a_i n_j\geq a_j n_i.
\]
For nonempty videos, this condition is equivalent to
\[
\frac{a_i}{a_i+n_i}\geq\frac{a_j}{a_j+n_j}.
\]
The pairwise preferences are thus induced by one scalar and are transitive. Sorting videos by anomaly
fraction maximizes every pairwise exchange and therefore the total number of concordant cross-video
pairs. Equal fractions can be tied or ordered arbitrarily without changing AUROC.

\section{Benchmark composition across VAD settings}
\label{app:benchmark-composition}

The decomposition in \Cref{eq:decomp} depends on frame labels and video
boundaries, not on the supervision used to train a detector.
\Cref{tab:vad-benchmark-composition} gives the numerical benchmark statistics
for \Cref{fig:vad-temporal-dilution} and records the common ShanghaiTech
subset used in Appendix~\ref{app:common-test}.
The table covers 13 benchmarks and distinguishes their test splits.
ShanghaiTech has separate 199-video weakly-supervised and 107-video
semi-supervised splits. MSAD Protocols II and I contain 240 and 360 test
videos, respectively.

\begin{table}[htbp]
\centering
\caption{Test-set composition across VAD settings. $V_{+}$ counts videos containing an anomaly. $w$ is the within-video pair fraction. Max. gain is the Micro-AUROC increase in percentage points from improving Within-AUROC from 50\% to 100\% while holding Cross-AUROC fixed. The final column gives the maximum Micro-AUROC (\%) with one anomaly score per video.}
\label{tab:vad-benchmark-composition}
\small
\setlength{\tabcolsep}{3pt}
\renewcommand{\arraystretch}{1.08}
\begin{tabular}{@{}lrrrrrr@{}}
\toprule
Benchmark & $V$ & $V_{+}$ & Frames & $w$ & \shortstack{Max. gain\\(pp)} & \shortstack{Maximum\\Micro-AUROC} \\
\midrule
\multicolumn{7}{@{}l}{\textit{Weakly-supervised test splits}} \\*
ShanghaiTech & 199 & 44 & 142,912 & $0.140\%$ & $0.070$ & $98.35$ \\
UCF-Crime & 290 & 140 & 1,114,144 & $0.388\%$ & $0.194$ & $92.95$ \\
XD-Violence & 800 & 500 & 2,330,384 & $0.071\%$ & $0.036$ & $95.46$ \\
MSAD (Protocol II) & 240 & 120 & 146,410 & $0.206\%$ & $0.103$ & $96.90$ \\
LAD2000 & 560 & 280 & 628,569 & $0.176\%$ & $0.088$ & $95.42$ \\
NTU CCTV-Fights & 250 & 250 & 438,345 & $0.526\%$ & $0.263$ & $78.92$ \\
VIEW360 & 200 & 105 & 153,641 & $0.401\%$ & $0.200$ & $88.21$ \\
\midrule
\multicolumn{7}{@{}l}{\textit{Semi-supervised test splits}} \\*
ShanghaiTech & 107 & 107 & 40,791 & $0.814\%$ & $0.407$ & $75.78$ \\
Avenue & 21 & 21 & 15,324 & $5.079\%$ & $2.540$ & $74.81$ \\
UCSD Ped1 & 36 & 36 & 7,200 & $2.267\%$ & $1.134$ & $74.58$ \\
UCSD Ped2 & 12 & 12 & 2,010 & $6.802\%$ & $3.401$ & $83.96$ \\
UBnormal & 211 & 158 & 92,640 & $0.190\%$ & $0.095$ & $93.86$ \\
MSAD (Protocol I) & 360 & 240 & 227,524 & $0.248\%$ & $0.124$ & $94.07$ \\
Street Scene & 35 & 35 & 146,410 & $1.631\%$ & $0.816$ & $88.42$ \\
IITB Corridor & 150 & 140 & 181,567 & $0.517\%$ & $0.259$ & $86.90$ \\
\midrule
\multicolumn{7}{@{}l}{\textit{Common test subset}} \\*
ShanghaiTech (common44) & 44 & 44 & 17,472 & $1.891\%$ & $0.945$ & $78.21$ \\
\bottomrule
\end{tabular}
\end{table}

Within-video pairs account for 6.802\% of comparisons on UCSD Ped2 and
5.079\% on Avenue, compared with 0.190\% on UBnormal and 0.071\% on
XD-Violence.
Cross-video pairs therefore account for most comparisons in each listed
test split, although the within-video fraction varies substantially.
These differences also affect the maximum Micro-AUROC with one anomaly
score per video. On ShanghaiTech, this maximum is 98.35\% on the
weakly-supervised split and 75.78\% on the semi-supervised split.
Thus, the same metric structure need not permit equally high scores
under different test-set compositions.

The table uses the full annotations for each listed protocol.
Detector results use the frames for which their anomaly scores are available.
The corresponding frame coverage is described in Appendix~\ref{app:protocol}
and the notes accompanying the supplementary results.

\section{Experimental setup}
\label{app:protocol}

\subsection{Methods and training protocol}
\label{app:grid}

The main WSVAD experiments evaluate five methods 
including Sultani, RTFM, CLIP-TSA, BN-WVAD, and S3R~\citep{sultani2018real,rtfm,joo2023cliptsa,zhou2024batchnorm,s3r}
with CLIP and I3D features~\citep{radford2021learning,carreira2017quo}
on ShanghaiTech, UCF-Crime, XD-Violence, and MSAD~\citep{liu2018future,sultani2018real,wu2020not,msad2024}.
We use the official implementation of each method and follow the
hyperparameters reported in the corresponding paper.
When a setting is not specified in the paper, we use the default
provided by the official implementation.
Each method, feature, and benchmark combination is trained with three
random seeds, giving 30 runs per benchmark and 120 detector runs in total.
Performance tables report the mean over the three runs, while retention
is computed for each run before aggregation.

Detector metrics are computed on the frames for which anomaly scores are
available and use the corresponding official frame annotations.
Ground-truth-only analyses instead use the full benchmark annotations.
For MSAD, this gives 143,920 evaluated frames for detector results,
compared with 146,410 frames in the full protocol.
The video-level anomaly classifiers are evaluated with both CLIP and I3D
features on all four benchmarks, with three runs for each configuration.

\paragraph{Mean-pool classifier training.}
The mean-pool classifier is trained on the standard training pool with video labels only. The training
configuration uses BCE, Adam with learning rate $10^{-3}$, batch size 128, 2,000 updates, and
final-iterate reporting.
There is no frame-label input, validation split, checkpoint selection, or test-time fitting. Pooling
is performed before the head, so the model receives one vector per video and cannot recover segment
order.

\paragraph{Tie convention.}
All AUROCs use mid-ranks, so a tied positive--negative pair contributes $0.5$~\citep{song2026vlm}. 
An anomaly score that is constant within each video
gives every mixed-label video's per-video AUROC a value of $0.5$, so Macro-AUROC and Within-AUROC
are both $0.5$. Pure-label videos have no defined per-video AUROC, so Macro-AUROC excludes them.
They also contribute no pair to Within-AUROC. This equality provides a deterministic validation
check. We also verify that all frames in a video receive the same anomaly score
after video-mean replacement.

\paragraph{Released outputs.}
Released denotes author-provided checkpoints or scores, including checkpoints for which we run inference.
Metrics use the recorded features and official GT. Feature substitutions are identified explicitly:
MGFN on UCF-Crime uses released weights with substitute features and is not the original feature configuration. MSAD RTFM and MGFN use released checkpoints with official I3D features.

\section{Additional video-level anomaly classification results}
\label{app:video-classifier-results}

\begin{table}[t]
\caption{Ground-truth-based controls with one anomaly score per video. One-bit uses
the binary video label, and the optimum orders videos by their fraction of
anomalous frames. MSAD uses the full benchmark annotations here. On the
frames used for detector evaluation, its two values are 89.57 and 96.90.}
\label{tab:constant-protocol-controls}
\centering
\small
\setlength{\tabcolsep}{3pt}
\begin{adjustbox}{max width=\linewidth}
\begin{tabular}{lrrrr}
\toprule
Benchmark & Videos & Frames & One-bit AUROC & Constant optimum \\
\midrule
ShanghaiTech & 199 & 142912 & $96.32$ & $98.35$ \\
XD-Violence & 800 & 2330384 & $84.93$ & $95.46$ \\
UCF-Crime & 290 & 1114144 & $81.57$ & $92.95$ \\
MSAD & 240 & 146410 & $89.53$ & $96.90$ \\
\bottomrule
\end{tabular}
\end{adjustbox}
\end{table}

\paragraph{One-bit video-composition control.}
This analytic control assigns one to every frame of a video containing any anomaly
and zero to every frame of an all-normal video. If $N_-^{\mathrm{anom}}$ is the number of negative
frames inside anomaly-containing videos and $N_-$ is the total negative-frame count, every positive
has score one and
\[
\auroc_{\mathrm{binary}}=1-\frac{1}{2}\frac{N_-^{\mathrm{anom}}}{N_-}.
\]
This gives \OneBitMicroAurocSht, \OneBitMicroAurocXd, \OneBitMicroAurocUcf{}, and $89.53\%$ Micro-AUROC on
ShanghaiTech, XD-Violence, UCF-Crime, and MSAD protocol GT, respectively
(Table~\ref{tab:constant-protocol-controls}). It uses test video labels as an analytic composition
control, not as a trainable method.

The mean-pool classifier averages features before classification. The other
three classifiers process the feature sequence before predicting one
anomaly score for the video. The sequence-linear head maps a fixed 32-position sequence
to one logit. The bi-GRU and Transformer encode that sequence and pool to one logit. These classifiers can use the temporal order of the features, but each
assigns its predicted video-level anomaly score to every frame for evaluation.

\begin{table}[t]
\caption{Frame-level AP (\%) of video-level anomaly classifiers. Entries are means
with sample standard deviations over three training runs.}
\label{tab:video-constant-family-ap}
\centering
\small
\setlength{\tabcolsep}{3pt}
\begin{adjustbox}{max width=\linewidth}
\begin{tabular}{llrrrr}
\toprule
Benchmark & Features & Mean-pool & Seq.-linear & Bi-GRU & Transformer \\
\midrule
ShanghaiTech & CLIP & $50.39{\scriptscriptstyle\pm}0.01$ & $45.61{\scriptscriptstyle\pm}0.07$ & $53.94{\scriptscriptstyle\pm}0.45$ & $54.45{\scriptscriptstyle\pm}2.18$ \\
ShanghaiTech & I3D & $42.26{\scriptscriptstyle\pm}0.17$ & $36.62{\scriptscriptstyle\pm}0.10$ & $41.12{\scriptscriptstyle\pm}0.81$ & $34.80{\scriptscriptstyle\pm}2.19$ \\
XD-Violence & CLIP & $65.36{\scriptscriptstyle\pm}0.09$ & $70.73{\scriptscriptstyle\pm}0.04$ & $69.76{\scriptscriptstyle\pm}0.47$ & $57.03{\scriptscriptstyle\pm}6.44$ \\
XD-Violence & I3D & $67.24{\scriptscriptstyle\pm}0.04$ & $65.92{\scriptscriptstyle\pm}0.48$ & $69.68{\scriptscriptstyle\pm}2.23$ & $56.47{\scriptscriptstyle\pm}7.29$ \\
UCF-Crime & CLIP & $17.50{\scriptscriptstyle\pm}0.06$ & $20.23{\scriptscriptstyle\pm}0.11$ & $19.48{\scriptscriptstyle\pm}1.12$ & $20.35{\scriptscriptstyle\pm}1.83$ \\
UCF-Crime & I3D & $19.59{\scriptscriptstyle\pm}0.25$ & $21.19{\scriptscriptstyle\pm}0.14$ & $17.77{\scriptscriptstyle\pm}0.08$ & $19.18{\scriptscriptstyle\pm}1.98$ \\
MSAD & CLIP & $63.34{\scriptscriptstyle\pm}0.05$ & $65.31{\scriptscriptstyle\pm}0.22$ & $67.46{\scriptscriptstyle\pm}0.99$ & $63.78{\scriptscriptstyle\pm}1.46$ \\
MSAD & I3D & $64.31{\scriptscriptstyle\pm}0.15$ & $65.81{\scriptscriptstyle\pm}0.16$ & $62.78{\scriptscriptstyle\pm}1.19$ & $55.11{\scriptscriptstyle\pm}2.74$ \\
\bottomrule
\end{tabular}
\end{adjustbox}
\end{table}

Table~\ref{tab:video-classifiers} reports the Micro-AUROC of the video-level
anomaly classifiers.
Table~\ref{tab:video-constant-family-ap} reports standard AP on all four benchmarks in a separate table;
AP remains XD-Violence's official metric and does not admit the AUROC pair decomposition.

\section{Detailed detector results}
\label{app:detector-results}

This section reports the numerical results underlying
\Cref{sec:detector-evaluation}. The 120 detector training runs and the 13
results from released checkpoints or scores remain separate throughout.

\subsection{Micro-AUROC decomposition}

\Cref{tab:public-decomp} gives Micro-AUROC, Cross-AUROC, Within-AUROC,
and Macro-AUROC for the original predictions, together with Micro-AUROC
after video-mean replacement. It includes all 40 combinations of method,
features, and benchmark from the training experiments.
Semi-supervised and training-free AUROC results are reported in
Appendix~\ref{app:broader-vad} and are not included in these WSVAD
retention summaries.

\begingroup
\footnotesize
\setlength{\tabcolsep}{3pt}
\renewcommand{\arraystretch}{1.06}
\begin{longtable}{@{}lllrrrrr@{}}
\caption{Detector AUROC results (\%). Mean Micro is Micro-AUROC after
video-mean replacement. $\dagger$ denotes the mean over three training runs.
Training-free AUROC results are in \Cref{tab:training-free-auroc}.}
\label{tab:public-decomp}\label{tab:public-auc}\\
\toprule
Benchmark & Method & Features & Micro & Mean Micro & Cross & Within & Macro \\
\midrule
\endfirsthead
\multicolumn{8}{@{}l}{\tablename\ \thetable\ (continued)} \\
\toprule
Benchmark & Method & Features & Micro & Mean Micro & Cross & Within & Macro \\
\midrule
\endhead
\midrule
\multicolumn{8}{r@{}}{Continued on the next page} \\
\endfoot
\bottomrule
\endlastfoot
\multicolumn{8}{@{}l}{\textit{Three-run averages}} \\
ShanghaiTech & Sultani$^{\dagger}$ & CLIP & $97.24$ & $96.87$ & $97.29$ & $66.24$ & $66.33$ \\
ShanghaiTech & Sultani$^{\dagger}$ & I3D & $85.00$ & $86.16$ & $85.01$ & $78.43$ & $77.49$ \\
ShanghaiTech & RTFM$^{\dagger}$ & CLIP & $96.05$ & $96.32$ & $96.13$ & $44.99$ & $44.34$ \\
ShanghaiTech & RTFM$^{\dagger}$ & I3D & $93.14$ & $94.01$ & $93.17$ & $75.45$ & $75.07$ \\
ShanghaiTech & CLIP-TSA$^{\dagger}$ & CLIP & $97.74$ & $95.93$ & $97.76$ & $82.59$ & $81.85$ \\
ShanghaiTech & CLIP-TSA$^{\dagger}$ & I3D & $86.76$ & $88.98$ & $86.77$ & $80.06$ & $79.45$ \\
ShanghaiTech & BN-WVAD$^{\dagger}$ & CLIP & $96.61$ & $96.27$ & $96.66$ & $61.37$ & $61.82$ \\
ShanghaiTech & BN-WVAD$^{\dagger}$ & I3D & $92.74$ & $92.27$ & $92.77$ & $70.90$ & $72.28$ \\
ShanghaiTech & S3R$^{\dagger}$ & CLIP & $97.99$ & $95.65$ & $98.02$ & $77.21$ & $80.02$ \\
ShanghaiTech & S3R$^{\dagger}$ & I3D & $94.26$ & $93.41$ & $94.27$ & $80.92$ & $80.11$ \\
\addlinespace[2pt]
UCF-Crime & Sultani$^{\dagger}$ & CLIP & $82.98$ & $80.08$ & $83.04$ & $68.11$ & $66.47$ \\
UCF-Crime & Sultani$^{\dagger}$ & I3D & $80.08$ & $78.89$ & $80.17$ & $57.30$ & $67.46$ \\
UCF-Crime & RTFM$^{\dagger}$ & CLIP & $79.91$ & $75.63$ & $79.97$ & $64.33$ & $64.84$ \\
UCF-Crime & RTFM$^{\dagger}$ & I3D & $81.39$ & $79.48$ & $81.48$ & $58.83$ & $65.77$ \\
UCF-Crime & CLIP-TSA$^{\dagger}$ & CLIP & $84.03$ & $80.77$ & $84.10$ & $65.60$ & $68.25$ \\
UCF-Crime & CLIP-TSA$^{\dagger}$ & I3D & $70.58$ & $71.90$ & $70.67$ & $48.60$ & $54.49$ \\
UCF-Crime & BN-WVAD$^{\dagger}$ & CLIP & $82.84$ & $80.04$ & $82.89$ & $70.45$ & $68.69$ \\
UCF-Crime & BN-WVAD$^{\dagger}$ & I3D & $80.19$ & $78.16$ & $80.27$ & $59.51$ & $67.00$ \\
UCF-Crime & S3R$^{\dagger}$ & CLIP & $80.84$ & $78.73$ & $80.89$ & $67.23$ & $68.55$ \\
UCF-Crime & S3R$^{\dagger}$ & I3D & $79.73$ & $76.95$ & $79.82$ & $56.95$ & $63.90$ \\
\addlinespace[2pt]
XD-Violence & Sultani$^{\dagger}$ & CLIP & $89.72$ & $90.68$ & $89.73$ & $78.58$ & $73.02$ \\
XD-Violence & Sultani$^{\dagger}$ & I3D & $87.58$ & $89.79$ & $87.58$ & $79.11$ & $74.70$ \\
XD-Violence & RTFM$^{\dagger}$ & CLIP & $90.21$ & $89.86$ & $90.22$ & $76.97$ & $73.40$ \\
XD-Violence & RTFM$^{\dagger}$ & I3D & $88.56$ & $89.65$ & $88.57$ & $79.33$ & $76.74$ \\
XD-Violence & CLIP-TSA$^{\dagger}$ & CLIP & $89.59$ & $89.46$ & $89.60$ & $74.15$ & $73.35$ \\
XD-Violence & CLIP-TSA$^{\dagger}$ & I3D & $88.82$ & $88.83$ & $88.82$ & $79.33$ & $77.92$ \\
XD-Violence & BN-WVAD$^{\dagger}$ & CLIP & $90.60$ & $88.84$ & $90.61$ & $82.84$ & $75.49$ \\
XD-Violence & BN-WVAD$^{\dagger}$ & I3D & $88.65$ & $86.99$ & $88.65$ & $79.52$ & $74.43$ \\
XD-Violence & S3R$^{\dagger}$ & CLIP & $91.34$ & $89.74$ & $91.34$ & $82.61$ & $76.77$ \\
XD-Violence & S3R$^{\dagger}$ & I3D & $87.15$ & $89.02$ & $87.17$ & $58.69$ & $59.78$ \\
\addlinespace[2pt]
MSAD & Sultani$^{\dagger}$ & CLIP & $90.79$ & $91.07$ & $90.86$ & $59.46$ & $63.57$ \\
MSAD & Sultani$^{\dagger}$ & I3D & $86.05$ & $86.70$ & $86.08$ & $68.19$ & $73.28$ \\
MSAD & RTFM$^{\dagger}$ & CLIP & $87.30$ & $87.03$ & $87.36$ & $59.76$ & $65.51$ \\
MSAD & RTFM$^{\dagger}$ & I3D & $84.45$ & $84.30$ & $84.48$ & $69.56$ & $68.22$ \\
MSAD & CLIP-TSA$^{\dagger}$ & CLIP & $88.99$ & $89.66$ & $89.05$ & $57.54$ & $56.78$ \\
MSAD & CLIP-TSA$^{\dagger}$ & I3D & $79.50$ & $84.47$ & $79.54$ & $63.24$ & $58.14$ \\
MSAD & BN-WVAD$^{\dagger}$ & CLIP & $87.65$ & $86.94$ & $87.71$ & $59.41$ & $63.81$ \\
MSAD & BN-WVAD$^{\dagger}$ & I3D & $87.42$ & $86.30$ & $87.46$ & $72.34$ & $80.03$ \\
MSAD & S3R$^{\dagger}$ & CLIP & $86.75$ & $86.36$ & $86.81$ & $59.28$ & $69.20$ \\
MSAD & S3R$^{\dagger}$ & I3D & $83.19$ & $82.71$ & $83.21$ & $71.70$ & $71.83$ \\
\midrule
\multicolumn{8}{@{}l}{\textit{Released checkpoints and scores}} \\
ShanghaiTech & S3R & I3D & $97.39$ & $95.75$ & $97.42$ & $82.38$ & $82.29$ \\
\addlinespace[2pt]
UCF-Crime & S3R & I3D & $85.99$ & $82.20$ & $86.08$ & $62.97$ & $71.40$ \\
UCF-Crime & DSANet & CLIP & $89.44$ & $85.34$ & $89.50$ & $73.72$ & $70.01$ \\
UCF-Crime & MGFN & I3D & $83.44$ & $80.54$ & $83.51$ & $63.03$ & $64.36$ \\
UCF-Crime & UR-DMU & I3D & $86.96$ & $83.14$ & $87.03$ & $68.30$ & $74.17$ \\
UCF-Crime & VadCLIP & CLIP & $88.01$ & $84.44$ & $88.07$ & $72.93$ & $66.62$ \\
\addlinespace[2pt]
XD-Violence & BN-WVAD & I3D & $94.71$ & $92.24$ & $94.72$ & $86.36$ & $78.54$ \\
XD-Violence & DSANet & CLIP & $95.40$ & $92.29$ & $95.41$ & $85.48$ & $79.85$ \\
XD-Violence & MGFN & I3D & $92.88$ & $90.85$ & $92.89$ & $81.84$ & $75.43$ \\
XD-Violence & UR-DMU & I3D & $94.02$ & $91.67$ & $94.02$ & $85.44$ & $77.91$ \\
XD-Violence & VadCLIP & CLIP & $94.68$ & $90.84$ & $94.69$ & $84.26$ & $75.29$ \\
\addlinespace[2pt]
MSAD & RTFM & I3D & $86.88$ & $85.83$ & $86.91$ & $73.36$ & $75.98$ \\
MSAD & MGFN & I3D & $84.96$ & $80.88$ & $84.98$ & $79.14$ & $79.18$ \\
\end{longtable}
\endgroup

\subsection{Video-mean replacement}
\label{app:mean-replacement}

\Cref{tab:video-mean-replacement-details} extends the run-level summary in
\Cref{tab:video-mean-replacement} to the released results.
Retention is computed before any averaging across runs.
The median is 98.8\% over the 120 training runs and 91.4\% over the 13
released results. The minimum values are 78.1\% and 88.3\%, respectively.

\begin{table}[htbp]
  \centering
  \caption{Micro-AUROC retention after video-mean replacement.
  Each training run contributes one $R_{\mathrm{mean}}$ value as defined in
  \Cref{eq:video-mean-replacement-evaluation}. Results from released
  checkpoints and scores are summarized separately. The final two columns
  count results retaining at least $85\%$ and $90\%$, respectively.}
  \label{tab:video-mean-replacement-details}
  \small
  \setlength{\tabcolsep}{5pt}
  \begin{tabular*}{\linewidth}{@{\extracolsep{\fill}}lrrrrr@{}}
  \toprule
  Benchmark & $n$ & Median $R_{\mathrm{mean}}$ & Min. $R_{\mathrm{mean}}$
  & $R_{\mathrm{mean}}\geq85\%$ & $R_{\mathrm{mean}}\geq90\%$ \\
  \midrule
  \multicolumn{6}{@{}l}{\textit{Detector training runs}} \\
  ShanghaiTech & 30 & $99.3\%$ & $95.0\%$ & $30/30$ & $30/30$ \\
  UCF-Crime & 30 & $92.6\%$ & $78.1\%$ & $29/30$ & $26/30$ \\
  XD-Violence & 30 & $100.0\%$ & $94.3\%$ & $30/30$ & $30/30$ \\
  MSAD & 30 & $100.1\%$ & $95.5\%$ & $30/30$ & $30/30$ \\
  All & 120 & $98.8\%$ & $78.1\%$ & $119/120$ & $116/120$ \\
  \midrule
  \multicolumn{6}{@{}l}{\textit{Released checkpoints and scores}} \\
  ShanghaiTech & 1 & $96.5\%$ & $96.5\%$ & $1/1$ & $1/1$ \\
  UCF-Crime & 5 & $89.7\%$ & $89.5\%$ & $5/5$ & $2/5$ \\
  XD-Violence & 5 & $94.5\%$ & $91.4\%$ & $5/5$ & $5/5$ \\
  MSAD & 2 & $92.7\%$ & $88.3\%$ & $2/2$ & $1/2$ \\
  All & 13 & $91.4\%$ & $88.3\%$ & $13/13$ & $9/13$ \\
  \bottomrule
  \end{tabular*}
\end{table}

\subsection{Frame-level AP}
\label{app:frame-ap}

For frame-level AP, we subtract the anomalous-frame fraction rather than
50\% when calculating retention. Let $\pi$ denote this fraction and let
$s$ and $\tilde{s}$ denote the original and replaced anomaly scores.
On the $[0,1]$ scale, AP retention is
\[
\frac{\mathrm{AP}(\tilde{s})-\pi}{\mathrm{AP}(s)-\pi}.
\]
\Cref{tab:video-mean-replacement-ap} summarizes this ratio when the original
AP exceeds $\pi$. One UCF-Crime training run does not satisfy this condition,
so it remains in the run count but is excluded from the ratio summary.

Across the training runs, median AP retention is 80.8\% on ShanghaiTech,
72.1\% on UCF-Crime, 94.6\% on XD-Violence, and 93.0\% on MSAD.
These results show that the response of frame-level AP depends on the
benchmark. They are empirical comparisons rather than consequences of
the Micro-AUROC decomposition.

\begin{table}[H]
  \centering
  \caption{Frame-level AP retention after video-mean replacement.
  $n$ counts evaluated results, and Defined $n$ counts those whose original
  AP exceeds the anomalous-frame fraction. The final three columns count
  increases in AP and retention of at least $85\%$ and $90\%$, respectively.}
  \label{tab:video-mean-replacement-ap}
  \small
  \setlength{\tabcolsep}{3pt}
  \begin{adjustbox}{max width=\linewidth}
  \begin{tabular}{llrrrrrrr}
  \toprule
  Benchmark & Source & $n$ & Defined $n$ & Median & Minimum
  & AP increases & Retention $\geq85\%$ & Retention $\geq90\%$ \\
  \midrule
  ShanghaiTech & Training runs & 30 & 30 & $80.8\%$ & $54.9\%$ & $3/30$ & $11/30$ & $9/30$ \\
  ShanghaiTech & Released & 1 & 1 & $67.3\%$ & $67.3\%$ & $0/1$ & $0/1$ & $0/1$ \\
  UCF-Crime & Training runs & 30 & 29 & $72.1\%$ & $50.0\%$ & $1/30$ & $5/29$ & $2/29$ \\
  UCF-Crime & Released & 5 & 5 & $63.9\%$ & $54.2\%$ & $0/5$ & $0/5$ & $0/5$ \\
  XD-Violence & Training runs & 30 & 30 & $94.6\%$ & $82.6\%$ & $6/30$ & $25/30$ & $22/30$ \\
  XD-Violence & Released & 5 & 5 & $85.6\%$ & $81.9\%$ & $0/5$ & $4/5$ & $0/5$ \\
  MSAD & Training runs & 30 & 30 & $93.0\%$ & $82.5\%$ & $3/30$ & $26/30$ & $21/30$ \\
  MSAD & Released & 2 & 2 & $87.9\%$ & $85.2\%$ & $0/2$ & $2/2$ & $1/2$ \\
  \bottomrule
  \end{tabular}
  \end{adjustbox}
\end{table}

\Cref{tab:public-ap} reports the original and replaced AP values for every
WSVAD method and feature configuration. The training-run rows average
AP over three runs. The retention summaries in
\Cref{tab:video-mean-replacement-ap} instead use the ratio calculated
separately for each run.

\begingroup
\footnotesize
\setlength{\tabcolsep}{3pt}
\renewcommand{\arraystretch}{1.06}
\begin{longtable}{@{}lllrr@{}}
\caption{Frame-level AP (\%) before and after video-mean replacement.
$\dagger$ denotes the mean over three training runs.
These are the same WSVAD predictions as in
\Cref{tab:public-decomp}.}
\label{tab:public-ap}\\
\toprule
Benchmark & Method & Features & Original AP & Video mean AP \\
\midrule
\endfirsthead
\multicolumn{5}{@{}l}{\tablename\ \thetable\ (continued)} \\
\toprule
Benchmark & Method & Features & Original AP & Video mean AP \\
\midrule
\endhead
\midrule
\multicolumn{5}{r@{}}{Continued on the next page} \\
\endfoot
\bottomrule
\endlastfoot
\multicolumn{5}{@{}l}{\textit{Three-run detector averages}} \\
ShanghaiTech & Sultani$^{\dagger}$ & CLIP & $55.83$ & $50.69$ \\
ShanghaiTech & Sultani$^{\dagger}$ & I3D & $43.07$ & $40.23$ \\
ShanghaiTech & RTFM$^{\dagger}$ & CLIP & $45.48$ & $48.24$ \\
ShanghaiTech & RTFM$^{\dagger}$ & I3D & $47.34$ & $39.38$ \\
ShanghaiTech & CLIP-TSA$^{\dagger}$ & CLIP & $63.24$ & $42.55$ \\
ShanghaiTech & CLIP-TSA$^{\dagger}$ & I3D & $49.74$ & $38.84$ \\
ShanghaiTech & BN-WVAD$^{\dagger}$ & CLIP & $49.03$ & $45.82$ \\
ShanghaiTech & BN-WVAD$^{\dagger}$ & I3D & $51.42$ & $44.40$ \\
ShanghaiTech & S3R$^{\dagger}$ & CLIP & $65.61$ & $39.18$ \\
ShanghaiTech & S3R$^{\dagger}$ & I3D & $54.41$ & $37.30$ \\
\addlinespace[2pt]
UCF-Crime & Sultani$^{\dagger}$ & CLIP & $22.88$ & $17.99$ \\
UCF-Crime & Sultani$^{\dagger}$ & I3D & $20.88$ & $17.00$ \\
UCF-Crime & RTFM$^{\dagger}$ & CLIP & $21.19$ & $15.82$ \\
UCF-Crime & RTFM$^{\dagger}$ & I3D & $21.66$ & $17.70$ \\
UCF-Crime & CLIP-TSA$^{\dagger}$ & CLIP & $24.79$ & $18.80$ \\
UCF-Crime & CLIP-TSA$^{\dagger}$ & I3D & $15.07$ & $14.17$ \\
UCF-Crime & BN-WVAD$^{\dagger}$ & CLIP & $22.12$ & $18.80$ \\
UCF-Crime & BN-WVAD$^{\dagger}$ & I3D & $21.03$ & $19.26$ \\
UCF-Crime & S3R$^{\dagger}$ & CLIP & $20.15$ & $17.17$ \\
UCF-Crime & S3R$^{\dagger}$ & I3D & $24.36$ & $16.85$ \\
\addlinespace[2pt]
XD-Violence & Sultani$^{\dagger}$ & CLIP & $72.47$ & $69.76$ \\
XD-Violence & Sultani$^{\dagger}$ & I3D & $68.61$ & $67.68$ \\
XD-Violence & RTFM$^{\dagger}$ & CLIP & $65.26$ & $67.41$ \\
XD-Violence & RTFM$^{\dagger}$ & I3D & $67.81$ & $64.11$ \\
XD-Violence & CLIP-TSA$^{\dagger}$ & CLIP & $64.89$ & $64.10$ \\
XD-Violence & CLIP-TSA$^{\dagger}$ & I3D & $69.93$ & $62.57$ \\
XD-Violence & BN-WVAD$^{\dagger}$ & CLIP & $67.86$ & $63.71$ \\
XD-Violence & BN-WVAD$^{\dagger}$ & I3D & $69.45$ & $62.17$ \\
XD-Violence & S3R$^{\dagger}$ & CLIP & $66.85$ & $65.70$ \\
XD-Violence & S3R$^{\dagger}$ & I3D & $54.96$ & $63.58$ \\
\addlinespace[2pt]
MSAD & Sultani$^{\dagger}$ & CLIP & $70.74$ & $69.02$ \\
MSAD & Sultani$^{\dagger}$ & I3D & $67.58$ & $63.84$ \\
MSAD & RTFM$^{\dagger}$ & CLIP & $64.30$ & $62.02$ \\
MSAD & RTFM$^{\dagger}$ & I3D & $62.98$ & $59.85$ \\
MSAD & CLIP-TSA$^{\dagger}$ & CLIP & $66.97$ & $66.55$ \\
MSAD & CLIP-TSA$^{\dagger}$ & I3D & $57.19$ & $58.85$ \\
MSAD & BN-WVAD$^{\dagger}$ & CLIP & $65.31$ & $61.91$ \\
MSAD & BN-WVAD$^{\dagger}$ & I3D & $64.89$ & $58.91$ \\
MSAD & S3R$^{\dagger}$ & CLIP & $62.79$ & $59.16$ \\
MSAD & S3R$^{\dagger}$ & I3D & $60.79$ & $55.89$ \\
\midrule
\multicolumn{5}{@{}l}{\textit{Released WSVAD checkpoints and scores}} \\
ShanghaiTech & S3R & I3D & $59.10$ & $41.61$ \\
\addlinespace[2pt]
UCF-Crime & S3R & I3D & $31.50$ & $20.55$ \\
UCF-Crime & DSANet & CLIP & $37.37$ & $25.57$ \\
UCF-Crime & MGFN & I3D & $23.54$ & $18.75$ \\
UCF-Crime & UR-DMU & I3D & $31.21$ & $26.52$ \\
UCF-Crime & VadCLIP & CLIP & $33.55$ & $24.17$ \\
\addlinespace[2pt]
XD-Violence & BN-WVAD & I3D & $84.93$ & $76.13$ \\
XD-Violence & DSANet & CLIP & $86.96$ & $77.43$ \\
XD-Violence & MGFN & I3D & $80.12$ & $73.22$ \\
XD-Violence & UR-DMU & I3D & $81.65$ & $73.21$ \\
XD-Violence & VadCLIP & CLIP & $84.52$ & $73.37$ \\
\addlinespace[2pt]
MSAD & RTFM & I3D & $66.29$ & $62.24$ \\
MSAD & MGFN & I3D & $63.54$ & $57.54$ \\
\midrule
\multicolumn{5}{@{}l}{\textit{Supplementary training-free methods}} \\
XD-Violence & LAVAD & VLM & $62.05$ & $60.15$ \\
XD-Violence & VADTree & VLM & $67.95$ & $61.92$ \\
\end{longtable}
\endgroup

\section{Additional VAD Results}
\label{app:broader-vad}

This appendix reports supplementary semi-supervised and VLM-based
VAD results.
For each method, we evaluate the original and replaced anomaly
scores on the same frames and annotations.
These results remain separate from the main WSVAD retention
summaries.

\subsection{Semi-supervised anomaly detection}
\label{app:semi-supervised}

\Cref{tab:semi-supervised-auroc} reports results on Avenue, UCSD Ped2,
and the 107-video ShanghaiTech split used for semi-supervised evaluation.
The methods include frame-prediction, object, pose, and trajectory-based
approaches, with their original feature representations.
Accurate-Interpretable-VAD produces identical anomaly scores across its
three recorded runs.
MNAD-Pred excludes the first four frames of each test video, which slightly
changes its within-video pair fraction.

\begin{table}[htbp]
\centering
\caption{Semi-supervised detector AUROC (\%). Mean Micro is Micro-AUROC after video-mean replacement, and $w$ is the within-video pair fraction (\%). $\dagger$ denotes three-run means. For these rows, Micro and Mean Micro include sample standard deviations.}
\label{tab:semi-supervised-auroc}
\footnotesize
\setlength{\tabcolsep}{3pt}
\renewcommand{\arraystretch}{1.08}
\begin{tabular}{@{}lrrrrrr@{}}
\toprule
Method & $w$ & Micro & Mean Micro & Cross & Within & Macro \\
\midrule
\multicolumn{7}{@{}l}{\textit{ShanghaiTech (107-video test split)}} \\*
TrajVAD-P & $0.814$ & $88.59$ & $61.92$ & $88.57$ & $91.58$ & $91.45$ \\
TrajVAD-T & $0.814$ & $84.89$ & $63.98$ & $84.89$ & $84.98$ & $87.00$ \\
MoCoDAD & $0.814$ & $76.68$ & $58.77$ & $76.66$ & $79.32$ & $77.52$ \\
SPARTA & $0.814$ & $84.83$ & $56.57$ & $84.80$ & $87.72$ & $89.69$ \\
STG-NF & $0.814$ & $85.94$ & $56.64$ & $85.92$ & $87.63$ & $89.20$ \\
UniVAD & $0.814$ & $89.32$ & $61.30$ & $89.30$ & $92.22$ & $91.01$ \\
\midrule
\multicolumn{7}{@{}l}{\textit{Avenue (21-video test split)}} \\*
Accurate-Interpretable-VAD$^{\dagger}$ & $5.079$ & $93.33{\scriptscriptstyle\pm}0.00$ & $63.15{\scriptscriptstyle\pm}0.00$ & $93.19$ & $95.91$ & $95.85$ \\
Jigsaw-VAD$^{\dagger}$ & $5.079$ & $90.32{\scriptscriptstyle\pm}0.37$ & $65.27{\scriptscriptstyle\pm}1.04$ & $90.21$ & $92.40$ & $88.94$ \\
MNAD-Pred$^{\dagger}$ & $5.099$ & $86.95{\scriptscriptstyle\pm}1.45$ & $63.49{\scriptscriptstyle\pm}1.82$ & $86.81$ & $89.59$ & $87.87$ \\
Jigsaw-VAD & $5.079$ & $92.10$ & $63.75$ & $92.04$ & $93.19$ & $91.47$ \\
\midrule
\multicolumn{7}{@{}l}{\textit{UCSD Ped2 (12-video test split)}} \\*
Accurate-Interpretable-VAD$^{\dagger}$ & $6.802$ & $99.06{\scriptscriptstyle\pm}0.00$ & $48.51{\scriptscriptstyle\pm}0.00$ & $99.05$ & $99.14$ & $99.46$ \\
Jigsaw-VAD$^{\dagger}$ & $6.802$ & $86.41{\scriptscriptstyle\pm}1.57$ & $61.72{\scriptscriptstyle\pm}2.78$ & $85.96$ & $92.57$ & $95.33$ \\
MNAD-Pred$^{\dagger}$ & $6.870$ & $92.99{\scriptscriptstyle\pm}1.68$ & $61.26{\scriptscriptstyle\pm}3.61$ & $92.87$ & $94.59$ & $95.66$ \\
\bottomrule
\end{tabular}
\end{table}

Micro-AUROC remains numerically close to Cross-AUROC in these results.
For example, TrajVAD-P on ShanghaiTech has Micro-AUROC of 88.59\% and
Cross-AUROC of 88.57\%, with a within-video pair fraction of 0.814\%.
Its Within-AUROC is 91.58\%.
Cross-video comparisons can therefore dominate the pooled metric even
when a detector distinguishes anomalous from normal frames well within
individual videos.

Video-mean replacement has a different effect from that observed in our
main WSVAD experiments.
TrajVAD-P's Micro-AUROC decreases from 88.59\% to 61.92\%.
On Avenue, the three-run Jigsaw-VAD result decreases from 90.32\% to
65.27\%. On UCSD Ped2, Accurate-Interpretable-VAD decreases from 99.06\%
to 48.51\%.
These changes show that a small within-video pair fraction does not by
itself imply that Micro-AUROC will remain high after replacement.
The decomposition fixes the weight of each pair population, whereas
replacement also changes the cross-video comparisons.

\subsection{VLM-Based Anomaly Detection}
\label{app:training-free}

\Cref{tab:training-free-auroc} reports additional VLM-based results.
LAVAD and VADTree perform detection without task-specific training.
VERA instead learns guiding questions from video-level labels while keeping the VLM weights fixed~\citep{ye2025vera}.
We also evaluate two AnomalyRuler configurations.
AnomalyRuler-base applies keyword rules and exponential averaging
to frame descriptions.
AnomalyRuler-Full adds Mistral deduction and uses the
implementation's keyword fallback.

\begin{table}[htbp]
\centering
\caption{
Additional VLM-based detector AUROC (\%) on the evaluated test splits. 
Mean Micro is Micro-AUROC after video-mean replacement, and $w$ is the within-video pair fraction (\%).
Each row evaluates one set of anomaly scores.
}
\label{tab:training-free-auroc}
\small
\setlength{\tabcolsep}{4pt}
\renewcommand{\arraystretch}{1.08}
\begin{tabular}{@{}lrrrrrr@{}}
\toprule
Method & $w$ & Micro & Mean Micro & Cross & Within & Macro \\
\midrule
\multicolumn{7}{@{}l}{\textit{UCF-Crime}} \\*
LAVAD & $0.388$ & $80.28$ & $79.66$ & $80.36$ & $61.75$ & $57.85$ \\
VADTree & $0.388$ & $84.74$ & $82.10$ & $84.78$ & $74.17$ & $63.70$ \\
VERA & $0.388$ & $85.58$ & $80.82$ & $85.62$ & $73.46$ & $70.27$ \\
\midrule
\multicolumn{7}{@{}l}{\textit{XD-Violence}} \\*
LAVAD & $0.071$ & $85.40$ & $85.32$ & $85.42$ & $64.78$ & $59.94$ \\
VADTree & $0.071$ & $90.49$ & $88.62$ & $90.50$ & $80.80$ & $73.94$ \\
VERA & $0.071$ & $85.50$ & $87.39$ & $85.51$ & $71.20$ & $64.28$ \\
\midrule
\multicolumn{7}{@{}l}{\textit{ShanghaiTech (107-video test split)}} \\*
AnomalyRuler-base & $0.814$ & $84.53$ & $63.86$ & $84.53$ & $84.10$ & $84.80$ \\
AnomalyRuler-Full & $0.814$ & $84.69$ & $62.96$ & $84.67$ & $87.30$ & $87.38$ \\
\midrule
\multicolumn{7}{@{}l}{\textit{Avenue}} \\*
AnomalyRuler-base & $5.079$ & $70.85$ & $50.12$ & $70.53$ & $76.68$ & $78.07$ \\
AnomalyRuler-Full & $5.079$ & $61.96$ & $44.45$ & $61.82$ & $64.54$ & $72.12$ \\
\midrule
\multicolumn{7}{@{}l}{\textit{UCSD Ped2}} \\*
AnomalyRuler-base & $6.802$ & $92.54$ & $71.39$ & $92.42$ & $94.24$ & $94.61$ \\
AnomalyRuler-Full & $6.802$ & $94.61$ & $75.34$ & $94.61$ & $94.70$ & $95.10$ \\
\bottomrule
\end{tabular}
\end{table}

On XD-Violence, LAVAD changes from 85.40\% to 85.32\% Micro-AUROC after
replacement, and VADTree changes from 90.49\% to 88.62\%.
Thus, high Micro-AUROC after assigning one anomaly score to every frame
of a video also occurs in training-free results.
The effect is not uniform across methods and test splits.
For example, AnomalyRuler-Full on Avenue decreases from 61.96\% to
44.45\%. The same decomposition applies throughout, but it does not
predict a common retention level.

\subsection{Evaluation on common test videos}
\label{app:common-test}

The weakly-supervised and semi-supervised ShanghaiTech protocols use
different test sets.
\Cref{tab:common-test-auroc} compares predictions on a shared set of
44 videos and 17,472 frames, using the same frame annotations.
All 44 videos contain an anomaly, and within-video pairs account for
1.891\% of anomalous--normal comparisons.
This fixes the evaluation frames, but does not hold the models or their
training procedures constant.

\begin{table}[htbp]
\centering
\caption{AUROC (\%) on the same 44 ShanghaiTech videos and 17,472 frames. All rows have $w=1.891\%$. Mean Micro is Micro-AUROC after video-mean replacement. $\dagger$ denotes the mean over three runs.}
\label{tab:common-test-auroc}
\footnotesize
\setlength{\tabcolsep}{3pt}
\renewcommand{\arraystretch}{1.08}
\begin{tabular}{@{}llrrrrr@{}}
\toprule
Method & Features & Micro & Mean Micro & Cross & Within & Macro \\
\midrule
\multicolumn{7}{@{}l}{\textit{Weakly-supervised}} \\*
BN-WVAD$^{\dagger}$ & CLIP & $54.09$ & $49.55$ & $53.95$ & $61.03$ & $61.42$ \\
BN-WVAD$^{\dagger}$ & I3D & $56.84$ & $52.31$ & $56.58$ & $70.30$ & $71.35$ \\
CLIP-TSA$^{\dagger}$ & CLIP & $68.58$ & $45.22$ & $68.32$ & $82.13$ & $81.37$ \\
CLIP-TSA$^{\dagger}$ & I3D & $61.07$ & $44.68$ & $60.76$ & $76.96$ & $74.24$ \\
RTFM$^{\dagger}$ & CLIP & $46.85$ & $50.05$ & $46.88$ & $45.11$ & $44.16$ \\
RTFM$^{\dagger}$ & I3D & $56.96$ & $41.34$ & $56.61$ & $74.97$ & $74.45$ \\
Sultani$^{\dagger}$ & CLIP & $63.31$ & $57.62$ & $63.26$ & $65.84$ & $66.03$ \\
Sultani$^{\dagger}$ & I3D & $69.00$ & $52.62$ & $68.83$ & $77.90$ & $75.96$ \\
S3R & I3D & $68.22$ & $44.70$ & $67.96$ & $81.80$ & $81.98$ \\
\midrule
\multicolumn{7}{@{}l}{\textit{Semi-supervised}} \\*
TrajVAD-P & Trajectory + pose & $89.66$ & $63.16$ & $89.59$ & $93.17$ & $93.35$ \\
TrajVAD-T & Trajectory & $85.47$ & $65.82$ & $85.48$ & $84.92$ & $87.84$ \\
MoCoDAD & Pose & $79.50$ & $59.71$ & $79.41$ & $83.80$ & $83.01$ \\
SPARTA & Pose & $84.24$ & $58.13$ & $84.17$ & $87.60$ & $88.34$ \\
STG-NF & Pose & $85.07$ & $56.36$ & $85.00$ & $88.38$ & $89.01$ \\
UniVAD & CLIP + pose & $89.25$ & $62.65$ & $89.20$ & $91.63$ & $90.43$ \\
\midrule
\multicolumn{7}{@{}l}{\textit{Training-free}} \\*
AnomalyRuler-base & Descriptions & $84.16$ & $65.65$ & $84.12$ & $85.92$ & $85.83$ \\
AnomalyRuler-Full & Descriptions + Mistral & $84.81$ & $64.80$ & $84.75$ & $87.77$ & $86.97$ \\
\bottomrule
\end{tabular}
\end{table}

On the shared test set, CLIP-TSA with CLIP features has Micro-AUROC of
68.58\% and Cross-AUROC of 68.32\%. Its Within-AUROC is 82.13\%.
TrajVAD-P has Micro-AUROC of 89.66\%, Cross-AUROC of 89.59\%,
and Within-AUROC of 93.17\%.
Video-mean replacement reduces their Micro-AUROC to 45.22\% and
63.16\%, respectively.
The dominance of cross-video comparisons persists on the shared frames,
while the change under replacement depends on the predictions being
evaluated. The shared test subset is a separate evaluation and is not
included in the 120-run WSVAD retention summary.

\section{Within-video weighting and anomaly-only evaluation}
\label{app:within-concentration}

Equation~\ref{eq:decomp} requires pair weights $a_i n_i$. Let
$\alpha_i=a_i n_i/W$ for $i\in\mathcal{M}$ and define
\begin{equation}
\neff=\frac{1}{\sum_{i\in\mathcal{M}}\alpha_i^2}.
\end{equation}
This inverse-concentration count is the number of equally contributing videos that would produce the
same pair-mass concentration. It is not a count of statistically independent videos and does not
estimate the variance of Within-AUROC.

\begin{table}[t]
\caption{Within-pair concentration on protocol GT. $V_{\mathrm{mix}}$ counts mixed-label videos and $N_{\mathrm{eff}}=(\sum_i a_i n_i)^2/\sum_i(a_i n_i)^2$ is a concentration statistic, not an independent sample size. MSAD uses protocol GT (97 mixed videos); its detector support has 96 mixed videos and $N_{\mathrm{eff}}=8.17$.}
\label{tab:within-concentration}
\centering
\small
\setlength{\tabcolsep}{3pt}
\begin{adjustbox}{max width=\linewidth}
\begin{tabular}{lrrr}
\toprule
Benchmark & $V_{\mathrm{mix}}$ & $N_{\mathrm{eff}}$ & $N_{\mathrm{eff}}/V_{\mathrm{mix}}$ \\
\midrule
ShanghaiTech & 43 & $28.3$ & $65.9\%$ \\
XD-Violence & 475 & $43.7$ & $9.2\%$ \\
UCF-Crime & 140 & $18.7$ & $13.3\%$ \\
MSAD & 97 & $8.3$ & $8.6\%$ \\
\bottomrule
\end{tabular}
\end{adjustbox}
\end{table}

Macro-AUROC gives equal weight to each video containing anomalous and
normal frames. Within-AUROC instead weights each video by $a_i n_i$,
as required by \Cref{eq:decomp}. Across the 120 detector training runs,
the absolute difference between the two metrics averages 3.76 percentage
points and reaches 12.73 percentage points.
\Cref{tab:public-decomp} reports both metrics for each method and feature
combination.

\subsection{Anomaly-only pooling}

Removing all-normal videos changes the set of frame comparisons but does
not restrict the remaining pairs to the same video.
\Cref{tab:auca} reports the resulting pair fractions and the maximum
AUC$_{\mathrm A}$ when every video receives one anomaly score.

\begin{table}[t]
\caption{Anomaly-only pooling under benchmark protocol GT. Cross mass kept is the fraction of original cross-video pairs surviving removal of all-normal videos. The final column gives the maximum AUC$_{\mathrm A}$ with one anomaly score per video. MSAD uses the 146,410-frame protocol axis, not the truncated detector support.}
\label{tab:auca}
\centering
\small
\setlength{\tabcolsep}{3pt}
\begin{adjustbox}{max width=\linewidth}
\begin{tabular}{lrrrr}
\toprule
Benchmark & Videos kept & Within-video fraction: Micro $\to$ AUC$_{\mathrm A}$ & Cross mass kept & Constant optimum \\
\midrule
ShanghaiTech & $44/199$ & $0.14\%\to 1.90\%$ & $7.2\%$ & $77.68$ \\
XD-Violence & $500/800$ & $0.07\%\to 0.24\%$ & $30.1\%$ & $84.92$ \\
UCF-Crime & $140/290$ & $0.39\%\to 1.05\%$ & $36.6\%$ & $80.87$ \\
MSAD & $120/240$ & $0.21\%\to 0.98\%$ & $20.8\%$ & $85.20$ \\
\bottomrule
\end{tabular}
\end{adjustbox}
\end{table}

\section{Temporal Interval Evaluation}
\label{app:temporal-map}
\label{sec:temporal-detection}

Temporal detection evaluates the overlap between predicted anomaly
intervals and annotated events.
We report binary anomaly-event AP for scalar detector scores and
video-level classifiers, and class-wise mAP for VadCLIP and DSANet.
The primary evaluation includes all test videos.
We also report evaluations restricted to anomaly-containing videos.
At each temporal intersection-over-union (tIoU) threshold, binary AP
evaluates one anomaly class,
whereas class-wise mAP averages the AP values of the retained classes.
Average AP and average mAP refer to the mean over tIoU thresholds
$0.1, 0.2, 0.3, 0.4,$ and $0.5$.
Metrics are calculated for each run before averaging across runs.

\paragraph{Evaluation population and class-wise reproduction.}
VadCLIP describes its fine-grained evaluation as using only abnormal
videos~\citep{wu2024vadclip}.
The released-code evaluation that includes all test videos instead
reproduces its reported average mAP of $6.68\%$ on UCF-Crime and
$24.70\%$ on XD-Violence.
Our corresponding values are $6.6806\%$ and $24.6953\%$
(\Cref{tab:temporal-native}).
Restricting the evaluation to anomaly-containing videos gives
$6.9226\%$ and $18.1577\%$, respectively.
We therefore use all test videos for the primary binary evaluation
and keep the restricted population as a separate comparison.

\paragraph{Original scores and video-mean replacement.}
The binary comparison uses a scalar adaptation of VadCLIP's released
interval evaluation.
We evaluate the intervals obtained from the original frame scores
and repeat interval extraction after replacing each video's scores
by their mean.
The predictions and their labels remain aligned to the same
recorded video boundaries in both conditions.
All 120 WSVAD training runs and 13 released WSVAD results yield
zero intervals and zero AP after replacement.

The released VadCLIP proposal rule retains consecutive frames whose
scores strictly exceed a threshold derived from their within-video
range.
For a constant score sequence, the threshold equals that constant,
so no frame is selected.
The zero-proposal outcomes in the binary runs agree with this behavior.
Class-wise replacement independently averages each output class over
time, and also yields zero proposals in all VadCLIP and DSANet runs.

\paragraph{Explicit whole-video predictions.}
We separately submit one interval $[0,T_i)$ for every evaluated video,
where $T_i$ is its evaluated length.
For a detector, the interval confidence is its mean anomaly score
$\bar{s}_i$.
For a video-level classifier, the confidence is its single output.
These intervals are supplied directly to the event evaluator.
They are not proposals extracted after video-mean replacement.
Within each comparison, both interval constructions use the same
evaluated frames, labels, and video boundaries.
Normal videos have no anomaly event, so anomaly predictions on
these videos are false positives.

\begin{table}[t]
\centering
\caption{
\textbf{Interval extraction and whole-video predictions give different results.}
Binary anomaly-event AP (\%) on all test videos.
The first two rows in each block use interval extraction.
Whole-video rows submit one interval per video.
Detector results use released checkpoints, and mean-pool results
average three runs with the indicated features.
Average AP uses tIoU thresholds $0.1, 0.2, 0.3, 0.4,$ and $0.5$.
}
\label{tab:temporal-detection-summary}
\small
\setlength{\tabcolsep}{3pt}
\renewcommand{\arraystretch}{1.08}

\begin{tabular*}{\linewidth}{@{\extracolsep{\fill}}lrr@{}}
\toprule
\multirow{2}{*}{Prediction} & \multicolumn{2}{c}{Anomaly-event AP} \\
\cmidrule(l){2-3}
& Average & tIoU $0.5$ \\
\midrule
\multicolumn{3}{@{}l}{\textit{MSAD, RTFM, I3D features}} \\
Original scores               & 46.64 & 21.82 \\
Video-mean replacement        & 0.00 & 0.00 \\
Whole video, detector mean    & 61.12 & 35.69 \\
Whole video, mean-pool        & 63.04 & 36.33 \\
\midrule
\multicolumn{3}{@{}l}{\textit{XD-Violence, VadCLIP, CLIP features}} \\
Original scores               & 19.19 & 8.85 \\
Video-mean replacement        & 0.00 & 0.00 \\
Whole video, detector mean    & 14.73 & 6.69 \\
Whole video, mean-pool        & 12.49 & 5.49 \\
\bottomrule
\end{tabular*}
\end{table}

\paragraph{Representative comparisons.}
\Cref{tab:temporal-detection-summary} separates interval extraction
from explicitly submitted whole-video predictions.
Video-mean replacement yields no proposals and zero AP.
On MSAD, however, whole-video intervals scored by RTFM's video mean
achieve $61.12\%$ average AP, compared with $46.64\%$ for intervals
extracted from its original scores.
The mean-pool classifier achieves $63.04\%$ with the same whole-video
interval construction.
On XD-Violence, VadCLIP's whole-video intervals instead achieve
$14.73\%$, below $19.19\%$ for extracted intervals.
The zero AP after replacement and the positive AP of directly
submitted whole-video intervals reflect different interval constructions.

\paragraph{WSVAD results.}
\Cref{tab:temporal-map-replacement} gives the 40 WSVAD method,
feature, and benchmark configurations, averaged over three runs,
and the 13 released WSVAD results.
LAVAD and VADTree are reported in a separate panel.
After video-mean replacement, both reported AP quantities are zero
for every row, so the table shows the original-score and explicit
whole-video conditions.
Whole-video intervals have higher average AP in 29 of 30 MSAD
training runs, compared with 7 of 30 on ShanghaiTech, 6 of 30 on
UCF-Crime, and 8 of 30 on XD-Violence.
The comparison therefore depends on the detector and benchmark.

\begingroup
\footnotesize
\setlength{\tabcolsep}{3pt}
\renewcommand{\arraystretch}{1.06}
\begin{longtable}{@{}lllrrrr@{}}
\caption{
Binary anomaly-event AP (\%) on all test videos.
Original denotes interval extraction from the original scalar scores.
Whole video denotes one interval per video scored by the detector mean.
Video-mean replacement followed by extraction gives zero AP throughout.
$\dagger$ marks three-run means.
}\label{tab:temporal-map-replacement}\\
\toprule
& & & \multicolumn{2}{c}{Average AP} & \multicolumn{2}{c}{AP@0.5} \\
\cmidrule(lr){4-5}\cmidrule(l){6-7}
Benchmark & Method & Features & Original & Whole video & Original & Whole video \\
\midrule
\endfirsthead
\multicolumn{7}{@{}l}{\tablename\ \thetable\ (continued)} \\
\toprule
& & & \multicolumn{2}{c}{Average AP} & \multicolumn{2}{c}{AP@0.5} \\
\cmidrule(lr){4-5}\cmidrule(l){6-7}
Benchmark & Method & Features & Original & Whole video & Original & Whole video \\
\midrule
\endhead
\midrule
\multicolumn{7}{r@{}}{Continued on the next page} \\
\endfoot
\bottomrule
\endlastfoot
\multicolumn{7}{@{}l}{\textit{Three-run detector means}} \\
ShanghaiTech & Sultani$^{\dagger}$ & CLIP & 31.85 & 28.04 & 8.96 & 4.85 \\
ShanghaiTech & Sultani$^{\dagger}$ & I3D & 21.41 & 23.31 & 8.07 & 5.51 \\
ShanghaiTech & RTFM$^{\dagger}$ & CLIP & 23.70 & 27.89 & 7.51 & 6.71 \\
ShanghaiTech & RTFM$^{\dagger}$ & I3D & 27.79 & 23.21 & 8.45 & 4.26 \\
ShanghaiTech & CLIP-TSA$^{\dagger}$ & CLIP & 34.69 & 24.94 & 12.88 & 4.53 \\
ShanghaiTech & CLIP-TSA$^{\dagger}$ & I3D & 26.77 & 22.16 & 9.90 & 5.13 \\
ShanghaiTech & BN-WVAD$^{\dagger}$ & CLIP & 25.70 & 26.43 & 5.98 & 5.00 \\
ShanghaiTech & BN-WVAD$^{\dagger}$ & I3D & 28.72 & 24.84 & 10.26 & 6.27 \\
ShanghaiTech & S3R$^{\dagger}$ & CLIP & 38.10 & 23.34 & 17.91 & 4.41 \\
ShanghaiTech & S3R$^{\dagger}$ & I3D & 31.18 & 23.87 & 12.27 & 4.08 \\
\addlinespace[2pt]
UCF-Crime & Sultani$^{\dagger}$ & CLIP & 11.88 & 10.35 & 2.04 & 0.50 \\
UCF-Crime & Sultani$^{\dagger}$ & I3D & 7.92 & 9.44 & 1.24 & 0.45 \\
UCF-Crime & RTFM$^{\dagger}$ & CLIP & 13.83 & 9.79 & 0.85 & 0.69 \\
UCF-Crime & RTFM$^{\dagger}$ & I3D & 13.99 & 9.36 & 2.06 & 0.71 \\
UCF-Crime & CLIP-TSA$^{\dagger}$ & CLIP & 14.88 & 9.88 & 3.40 & 0.71 \\
UCF-Crime & CLIP-TSA$^{\dagger}$ & I3D & 8.13 & 9.60 & 1.28 & 0.41 \\
UCF-Crime & BN-WVAD$^{\dagger}$ & CLIP & 14.08 & 10.43 & 3.24 & 0.63 \\
UCF-Crime & BN-WVAD$^{\dagger}$ & I3D & 9.28 & 9.81 & 2.12 & 0.58 \\
UCF-Crime & S3R$^{\dagger}$ & CLIP & 11.15 & 9.55 & 2.23 & 0.54 \\
UCF-Crime & S3R$^{\dagger}$ & I3D & 12.75 & 8.84 & 1.14 & 0.67 \\
\addlinespace[2pt]
XD-Violence & Sultani$^{\dagger}$ & CLIP & 9.32 & 13.49 & 2.37 & 6.27 \\
XD-Violence & Sultani$^{\dagger}$ & I3D & 6.67 & 13.84 & 1.19 & 5.95 \\
XD-Violence & RTFM$^{\dagger}$ & CLIP & 19.45 & 12.66 & 8.08 & 5.41 \\
XD-Violence & RTFM$^{\dagger}$ & I3D & 24.57 & 12.79 & 9.29 & 5.20 \\
XD-Violence & CLIP-TSA$^{\dagger}$ & CLIP & 19.86 & 11.57 & 6.89 & 4.38 \\
XD-Violence & CLIP-TSA$^{\dagger}$ & I3D & 18.48 & 12.38 & 6.77 & 4.90 \\
XD-Violence & BN-WVAD$^{\dagger}$ & CLIP & 16.80 & 12.00 & 6.44 & 4.87 \\
XD-Violence & BN-WVAD$^{\dagger}$ & I3D & 15.55 & 13.23 & 5.39 & 5.15 \\
XD-Violence & S3R$^{\dagger}$ & CLIP & 21.70 & 12.28 & 7.89 & 5.10 \\
XD-Violence & S3R$^{\dagger}$ & I3D & 12.19 & 11.94 & 3.91 & 4.07 \\
\addlinespace[2pt]
MSAD & Sultani$^{\dagger}$ & CLIP & 32.72 & 67.99 & 17.96 & 38.20 \\
MSAD & Sultani$^{\dagger}$ & I3D & 24.30 & 63.85 & 10.80 & 36.87 \\
MSAD & RTFM$^{\dagger}$ & CLIP & 46.85 & 62.13 & 28.22 & 36.46 \\
MSAD & RTFM$^{\dagger}$ & I3D & 42.67 & 60.53 & 20.29 & 35.42 \\
MSAD & CLIP-TSA$^{\dagger}$ & CLIP & 40.58 & 65.92 & 18.74 & 36.10 \\
MSAD & CLIP-TSA$^{\dagger}$ & I3D & 39.47 & 60.04 & 18.76 & 32.27 \\
MSAD & BN-WVAD$^{\dagger}$ & CLIP & 46.10 & 65.92 & 24.86 & 38.00 \\
MSAD & BN-WVAD$^{\dagger}$ & I3D & 53.17 & 62.63 & 30.24 & 35.93 \\
MSAD & S3R$^{\dagger}$ & CLIP & 54.69 & 59.86 & 31.24 & 33.99 \\
MSAD & S3R$^{\dagger}$ & I3D & 44.73 & 58.48 & 17.76 & 33.24 \\
\addlinespace[2pt]
\midrule
\multicolumn{7}{@{}l}{\textit{Released WSVAD checkpoints and scores}} \\
ShanghaiTech & S3R & I3D & 34.39 & 25.24 & 10.85 & 5.22 \\
\addlinespace[2pt]
UCF-Crime & DSANet & CLIP & 19.08 & 10.98 & 6.04 & 0.82 \\
UCF-Crime & MGFN & I3D & 13.06 & 9.77 & 1.92 & 1.04 \\
UCF-Crime & S3R & I3D & 14.57 & 9.56 & 2.48 & 0.83 \\
UCF-Crime & UR-DMU & I3D & 18.72 & 10.86 & 6.31 & 0.79 \\
UCF-Crime & VadCLIP & CLIP & 14.55 & 11.71 & 2.84 & 0.73 \\
\addlinespace[2pt]
XD-Violence & BN-WVAD & I3D & 17.36 & 15.42 & 8.84 & 6.86 \\
XD-Violence & DSANet & CLIP & 21.53 & 15.24 & 10.29 & 7.26 \\
XD-Violence & MGFN & I3D & 22.82 & 14.17 & 10.58 & 6.85 \\
XD-Violence & UR-DMU & I3D & 26.00 & 14.13 & 11.07 & 6.48 \\
XD-Violence & VadCLIP & CLIP & 19.19 & 14.73 & 8.85 & 6.69 \\
\addlinespace[2pt]
MSAD & MGFN & I3D & 61.75 & 56.25 & 39.99 & 33.51 \\
MSAD & RTFM & I3D & 46.64 & 61.12 & 21.82 & 35.69 \\
\addlinespace[2pt]
\midrule
\multicolumn{7}{@{}l}{\textit{Additional training-free results}} \\
UCF-Crime & LAVAD & VLM & 3.83 & 9.84 & 0.75 & 1.01 \\
UCF-Crime & VADTree & VLM & 18.13 & 13.43 & 4.85 & 1.50 \\
XD-Violence & LAVAD & VLM & 4.66 & 13.25 & 0.87 & 4.93 \\
XD-Violence & VADTree & VLM & 16.48 & 13.22 & 6.68 & 4.69 \\
\end{longtable}
\endgroup

\paragraph{Video-level classifiers.}
\Cref{tab:temporal-video-classifiers} reports all four classifier
architectures with both feature representations.
Each value is a mean of three separately evaluated training runs.
Every prediction is a whole-video interval, so the classifiers do
not predict event boundaries within a video.
The table reports average AP and AP@0.5 on all test videos.

\begin{table}[!ht]
\centering
\caption{
Binary anomaly-event AP (\%) for whole-video intervals scored by
video-level classifiers. Each entry averages three runs.
The interval construction is identical for every classifier.
}
\label{tab:temporal-video-classifiers}
\footnotesize
\setlength{\tabcolsep}{4pt}
\renewcommand{\arraystretch}{1.06}
\begin{tabular*}{\linewidth}{@{\extracolsep{\fill}}llrrrr@{}}
\toprule
Classifier & Features & ShanghaiTech & UCF-Crime & XD-Violence & MSAD \\
\midrule
\multicolumn{6}{@{}l}{\textit{All test videos, Average AP}} \\
Mean-pool & CLIP & 28.11 & 9.99 & 12.49 & 66.44 \\
Sequence-linear & CLIP & 27.13 & 11.11 & 14.17 & 67.04 \\
Bi-GRU & CLIP & 29.17 & 10.92 & 13.31 & 66.40 \\
Transformer & CLIP & 29.51 & 11.08 & 11.28 & 66.37 \\
Mean-pool & I3D & 23.28 & 9.43 & 13.89 & 63.04 \\
Sequence-linear & I3D & 23.37 & 9.90 & 13.65 & 63.87 \\
Bi-GRU & I3D & 22.51 & 9.35 & 13.16 & 61.73 \\
Transformer & I3D & 21.56 & 10.29 & 11.54 & 61.12 \\
\midrule
\multicolumn{6}{@{}l}{\textit{All test videos, AP@0.5}} \\
Mean-pool & CLIP & 5.80 & 0.48 & 5.49 & 36.01 \\
Sequence-linear & CLIP & 4.85 & 0.66 & 6.61 & 36.77 \\
Bi-GRU & CLIP & 5.60 & 0.62 & 5.87 & 37.17 \\
Transformer & CLIP & 6.40 & 0.45 & 3.86 & 36.08 \\
Mean-pool & I3D & 5.56 & 0.57 & 5.90 & 36.33 \\
Sequence-linear & I3D & 4.62 & 0.64 & 5.43 & 37.08 \\
Bi-GRU & I3D & 7.21 & 0.62 & 4.97 & 35.74 \\
Transformer & I3D & 3.65 & 0.59 & 3.52 & 32.21 \\
\bottomrule
\end{tabular*}
\end{table}

\paragraph{Anomaly-containing videos.}
Restricting the binary evaluation to anomaly-containing videos
preserves the contrast between the RTFM and VadCLIP examples.
On MSAD, RTFM's whole-video intervals achieve $69.71\%$ average AP,
compared with $51.49\%$ for extracted intervals.
On XD-Violence, VadCLIP's whole-video intervals achieve $14.78\%$,
below $19.27\%$ for extracted intervals.
Video-mean replacement still yields zero AP for every detector.
The separate class-wise comparisons are reported in
\Cref{tab:temporal-native}.

\begin{table}[t]
\centering
\caption{
Class-wise mAP (\%) from released VadCLIP and DSANet checkpoints.
Each population retains the classes recorded by the released evaluation,
14 on UCF-Crime and 7 on XD-Violence.
Every class-wise video-mean-replacement run produces zero proposals
and zero mAP at all reported thresholds.
}
\label{tab:temporal-native}
\small
\setlength{\tabcolsep}{4pt}
\begin{tabular*}{\linewidth}{@{\extracolsep{\fill}}llrrrr@{}}
\toprule
& & \multicolumn{2}{c}{All test videos}
& \multicolumn{2}{c}{Anomaly-containing videos} \\
\cmidrule(lr){3-4}\cmidrule(l){5-6}
Benchmark & Method & Average & mAP@0.5 & Average & mAP@0.5 \\
\midrule
UCF-Crime & VadCLIP & 6.68 & 2.93 & 6.92 & 3.04 \\
UCF-Crime & DSANet & 13.01 & 8.00 & 13.17 & 8.06 \\
XD-Violence & VadCLIP & 24.70 & 14.34 & 18.16 & 9.66 \\
XD-Violence & DSANet & 28.87 & 17.89 & 18.88 & 9.45 \\
\bottomrule
\end{tabular*}
\end{table}

\paragraph{Stricter temporal overlap.}
The binary runs additionally report AP at tIoU $0.7$ and $0.9$.
For released RTFM on MSAD, the original-score results are $11.71\%$
and $1.99\%$, compared with $20.82\%$ and $8.68\%$ for whole-video
intervals.
A whole-video interval has tIoU equal to the length of an individual
annotated event divided by the evaluated video length.
It can match at most one event, so short or multiple events constrain
this baseline independently of its confidence ordering.

\section{Metric reporting across VAD training settings}
\label{app:metric-survey}

We surveyed 113 peer-reviewed VAD papers published from 2018 through 2026 that report a
numerical result for their own method on at least one benchmark in
Table~\ref{tab:vad-benchmark-composition}. 
The 59 weakly supervised methods use target-domain
video-level anomaly labels.
This group includes methods with category-level video labels,
few-shot supervision, or continual updates.
The 45 semi-supervised methods learn normality from
normal target videos. They may use pretrained feature extractors or synthetic perturbations of
normal data, but do not train on real labeled anomaly clips. The nine training-free methods do
not fit VAD detector weights to the target benchmark. Some use normal references, update an
inference memory, or select score-combination parameters using benchmark results. These
differences in available information matter when comparing performance across groups.

We count a metric when a paper gives a numerical result for its proposed method, including in
ablations or analysis. An
unspecified frame-level AUC counts as frame-level reporting without assuming that the authors
pooled all frames to compute Micro-AUROC. AUC$_{\mathrm A}$ requires one score over all test
videos containing anomalies, rather than separate scores for anomaly categories. Macro-AUROC
requires an explicit average of per-video AUROCs. Event-level mAP requires temporal overlap
between a predicted interval and a ground-truth event. For FAR, we count numerical values
reported by the authors, while recording differences in evaluation conditions below.
Table~\ref{tab:metric-survey} summarizes these reports.

\begin{table}[h]
\caption{
\textbf{Frame-level AUROC or AP dominates VAD evaluation.}
Values show the percentage of surveyed papers reporting each metric,
with paper counts in parentheses.
Only papers reporting their own numerical results on a benchmark in
Table~\ref{tab:vad-benchmark-composition} are counted.
AUC$_{\mathrm A}$ pools frames from all videos containing anomalies.
FAR follows the definition reported in each paper.
}
\label{tab:metric-survey}
\begin{center}
\setlength{\tabcolsep}{6pt}
\begin{tabular}{@{}lrrrr@{}}
\toprule
Metric
& WSVAD
& Semi-supervised
& Training-free
& Overall \\
& $(n=59)$
& $(n=45)$
& $(n=9)$
& $(n=113)$ \\
\midrule
Frame-level AUROC or AP
& $98.3\%$ $(58)$
& $100\%$ $(45)$
& $100\%$ $(9)$
& $\mathbf{99.1\%}$ $(112)$ \\
\midrule
AUC$_{\mathrm A}$
& $18.6\%$ $(11)$
& $0\%$ $(0)$
& $0\%$ $(0)$
& $9.7\%$ $(11)$ \\
Macro-AUROC
& $0\%$ $(0)$
& $22.2\%$ $(10)$
& $0\%$ $(0)$
& $8.8\%$ $(10)$ \\
Event-level mAP at temporal IoU
& $13.6\%$ $(8)$
& $0\%$ $(0)$
& $0\%$ $(0)$
& $7.1\%$ $(8)$ \\
Reported FAR
& $13.6\%$ $(8)$
& $0\%$ $(0)$
& $0\%$ $(0)$
& $7.1\%$ $(8)$ \\
\bottomrule
\end{tabular}
\end{center}
\end{table}

The weakly supervised papers published from 2018 through 2022 are
\citep{sultani2018real,zhong2019graph,wu2020not,zaheer2020claws,rtfm,lv2021localizing,feng2021mist,msl,s3r,wu2022weakly,yu2022modality,panariello2022consistency}.
The 2023 and 2024 papers are
\citep{zhou2023dual,mgfn,lv2023unbiased,zhang2023exploiting,joo2023cliptsa,pang2023audiovisual,zhou2024batchnorm,wu2024vadclip,chen2024prompt,karim2024realtime,tan2024overlooked,ghadiya2024crossmodal,wu2024weakly,jain2024crossdomain,cho2024towards,leng2024beyond,sun2024multimodal,sun2024eventdriven,lee2024cognitive,yang2024normality,pu2024prompt,wu2024openvocab}.
The 2025 and 2026 papers are
\citep{song2025anomaly,damicantonio2025mixture,ye2025vera,qiu2025learning,jin2025aligning,li2025wsvadclip,meng2025audiovisual,chu2025scene,dalvi2025distilling,xu2025suppression,qiu2025temporal,he2025spmil,xu2026tlma,acharya2026road,zhao2026learning,sun2026mome,zhang2026fine,yin2026learning,chu2026reba,sun2026enhancing,yang2026eks,wang2026las,senapati2026poinclip,zheng2026logical,hashimoto2026cade}.
Open-vocabulary, few-shot, and continual evaluations appear in this group
\citep{wu2024openvocab,zheng2026logical,hashimoto2026cade}. They retain
video-level supervision but use protocols that differ from conventional binary WSVAD.
\citet{song2025anomaly} also use direction labels for an auxiliary task on VIEW360.

The semi-supervised papers published from 2018 through 2022 are
\citep{liu2018future,morais2019regularity,gong2019memorizing,ionescu2019object,nguyen2019appearance,park2020memory,yu2020cloze,lu2020fewshot,georgescu2021multitask,astrid2021synthetic,astrid2021learning,luo2021normalgraph,szymanowicz2021xman,wu2022explainable,wang2022video,ristea2022sspcab,park2022fastano,cho2022flows}.
The 2023 and 2024 papers are
\citep{barbalau2023ssmtl,cao2023benchmark,yang2023restoration,astrid2023pseudobound,yan2023diffusion,shi2023pretext,baradaran2023attention,wang2023ensemble,zhang2024multiscale,micorek2024mulde,zeng2024progressive,cheng2024dhvad,shao2024consistency,liu2024vadiffusion,li2024vlavad,lappas2024distinction,rai2024pseudo,msad2024}.
The 2025 and 2026 papers are
\citep{jiang2025local,zhang2025autoregressive,reiss2025attribute,zhou2025mapdm,hu2025noiseresistant,yu2026stepdiffusion,li2026causal,mumcu2026mllm,song2026bounding}.
Macro-AUROC is reported by
\citep{wang2022video,ristea2022sspcab,barbalau2023ssmtl,zhang2024multiscale,micorek2024mulde,zeng2024progressive,msad2024,jiang2025local,reiss2025attribute,zhang2025autoregressive}.
The MSAD study reports Macro-AUROC under cross-view and cross-scenario tests, and
\citet{mumcu2026mllm} report an Avenue result for their method combined with T-EVAL.
Several other semi-supervised studies use externally pretrained object or pose models, normal
support sets, or synthetic anomalies. Their frame-level AUROC reports still concern normal-only VAD
training, but their data and adaptation protocols are not identical.

The training-free papers are
\citep{lavad,yang2024followrules,ahn2024videopatchcore,shao2025eventvad,li2026vadtree,ahn2026anyanomaly,feng2026monitor,lin2025holistic,mo2026reci}.
Normal references are used by \citet{yang2024followrules,ahn2024videopatchcore}.
\citet{feng2026monitor,mo2026reci} update their inference context as a video stream is
processed. \citet{ahn2026anyanomaly,lin2025holistic} select score parameters using benchmark
results. Thus, training-free here describes detector-weight fitting, not the absence of
reference data or score calibration.

Among weakly supervised papers, AUC$_{\mathrm A}$ is reported by
\citep{lv2021localizing,zhou2023dual,lv2023unbiased,zhou2024batchnorm,wu2024vadclip,damicantonio2025mixture,qiu2025learning,xu2025suppression,xu2026tlma,acharya2026road,chu2026reba}.
Temporal event-level mAP appears in
\citep{wu2022weakly,wu2024vadclip,li2025wsvadclip,zhang2026fine,yin2026learning,chu2026reba,wang2026las,senapati2026poinclip}.
Numerical FAR results appear in
\citep{sultani2018real,zhong2019graph,feng2021mist,wu2022weakly,zhou2023dual,pang2023audiovisual,pu2024prompt,xu2026tlma}.
\citet{zhou2023dual} do not specify the score threshold for their FAR results.
\citet{feng2021mist} also omit the threshold and describe their FAR as measured on anomaly
videos. These reporting differences limit direct comparisons of FAR values.

The metric names need a few qualifications. \citet{zhang2026fine} report temporal event
mAP but no frame-level AUROC or AP.
\citet{shao2025eventvad} report frame-level AUROC and AP,
but not temporal event-level mAP on the included benchmarks. The category-specific
anomaly-only AUCs of \citet{pang2023audiovisual} are not one pooled
AUC$_{\mathrm A}$. Most semi-supervised test splits in
Table~\ref{tab:vad-benchmark-composition} contain no all-normal videos. On those splits,
overall pooled AUROC already uses the same test videos as AUC$_{\mathrm A}$, even when the paper
does not name it separately.

Frame-level AUROC or AP is reported in 112 of the 113 surveyed papers across weakly supervised, semi-supervised, and training-free VAD.
Alternative evaluations are substantially less common.
Macro-AUROC appears in 10 papers, all in the semi-supervised setting, while temporal event-level mAP appears in 8 weakly supervised papers.
These reporting patterns motivate separating the different questions that frame-level pooled ranking, within-video ranking, and interval-based evaluation answer.

\end{document}